\documentclass{article}
\usepackage{graphicx}
\usepackage{amsmath} 
\usepackage{multirow}
\usepackage[document]{ragged2e}
\usepackage{cite}
\usepackage[utf8]{inputenc}
\usepackage{appendix}
\usepackage{xcolor}
\usepackage{todonotes}
\usepackage[preprint, nonatbib]{neurips_2023}
\usepackage[utf8]{inputenc} % allow utf-8 input
\usepackage[T1]{fontenc}    % use 8-bit T1 fonts
\usepackage{hyperref}       % hyperlinks
\usepackage{url}            % simple URL typesetting
\usepackage{booktabs}       % professional-quality tables
\usepackage{amsfonts}       % blackboard math symbols
\usepackage{nicefrac}       % compact symbols for 1/2, etc.
\usepackage{microtype}      % microtypography
\usepackage{xcolor}         % colors
\usepackage{ragged2e}
\usepackage{enumitem}
\usepackage{mathtools}

\title{
NoisyNN: Exploring the Impact of Information Entropy Change in Learning Systems}

\author{%
  Xiaowei Yu\footnotemark[1]\\
  CS \\
  Missouri S\&T \\
  \And
  Zhe Huang\footnotemark[1]\\
  CS\\
  Tufts University \\
 \And
  Minheng Chen\\
  CSE\\
  UT-Arlington \\
\And
  Lu Zhang\\
  CSE\\
  UT-Arlington \\
  \And
  Li Wang\\
  Department of Mathematics\\
  UT-Arlington \\
  \And
  Tianming Liu\\
  School of Computing\\
  University of Georgia\\
  \And
  Dajiang Zhu\\
  CSE\\
  UT-Arlington \\
}

\begin{document}

\maketitle
\renewcommand{\thefootnote}{\fnsymbol{footnote}} %将脚注符号设置为fnsymbol类型，即特殊符号表示
\footnotetext[1]{Equal contribution.} %对应脚注[1]

\justifying{
\begin{abstract}
We explore the impact of entropy change in deep learning systems via noise injection at different levels, i.e., embeddings and images. The series of models that employ our methodology are collectively known as Noisy Neural Networks (NoisyNN), with examples such as NoisyViT and NoisyCNN discussed in the paper. Noise is conventionally viewed as a harmful perturbation in various deep learning models, such as convolutional neural networks (CNNs) and vision transformers (ViTs), as well as different learning tasks. However, this work shows that noise can be an effective way to change the entropy of the learning task. We demonstrate that specific noise can boost the performance of various deep models under certain conditions. We theoretically prove the enhancement gained from positive noise by reducing the task complexity defined by information entropy and experimentally show the significant performance gain in large image datasets, such as ImageNet. Herein, we use the information entropy to define the complexity of the task. We categorize the noise into two types, positive noise (PN) and harmful noise (HN), based on whether the noise can help reduce the complexity of the task. Extensive experiments of CNNs and ViTs have shown performance improvements by proactively injecting positive noise, where we achieved an unprecedented top 1 accuracy of over 95$\%$ on ImageNet. Both theoretical analysis and empirical evidence have confirmed that the presence of positive noise, can benefit the learning process, while the traditionally perceived harmful noise indeed impairs deep learning models. The different roles of noise offer new explanations for deep models on specific tasks and provide a new paradigm for improving model performance. Moreover, it reminds us that we can influence the performance of learning systems via information entropy change.
\end{abstract}
}
%The theoretical analysis also gives directions to implement effective and efficient data augmentation works in deep learning. 

\setlength{\parindent}{0pt}
\section{Introduction}
Noise, conventionally regarded as a hurdle in machine learning and deep learning tasks, is universal and unavoidable due to various reasons, e.g., environmental factors, instrumental calibration, and human activities \cite{Ormiston20} \cite{Thulasidasan19}. In computer vision, noise can be generated from different phases: (1) Image Acquisition: Noise can arise from a camera sensor or other imaging device \cite{Sijbers1996}. For example, electronic or thermal noise in the camera sensor can result in random pixel values or color variations that can be visible in the captured image. (2) Image Preprocessing: Noise can be introduced during preprocessing steps such as image resizing, filtering, or color space conversion \cite{Shaykh98}. For example, resizing an image can introduce aliasing artifacts, while filtering an image can result in the loss of detail and texture. (3) Feature Extraction: Feature extraction algorithms can be sensitive to noise in the input image, which can result in inaccurate or inconsistent feature representations \cite{Albukhanajer2014}. For example, edge detection algorithms can be affected by noise in the image, resulting in false positives or negatives. (4) Algorithms: algorithms used for computer vision tasks, such as object detection or image segmentation, can also be sensitive to noise in the input data \cite{Braun2021}. Noise can cause the algorithm to learn incorrect patterns or features, leading to poor performance. 
%Several examples of the influence of noise on the image are provided in Fig. \ref{noiseBG}.

Since noise is an unavoidable reality in engineering tasks, existing works usually make the assumption that noise has a consistently negative impact on the current task \cite{Sethna2001} \cite{Owotogbe19}. Nevertheless, is the above assumption always valid? As such, it is crucial to address the question of whether noise can ever have a positive influence on deep learning models. This work aims to provide a comprehensive answer to this question, which is a pressing concern in the deep learning community. We recognize that the imprecise definition of noise is a critical factor leading to the uncertainties surrounding the identification and characterization of noise. To address these uncertainties, an in-depth analysis of the task's complexity is imperative for arriving at a rigorous answer. By using the definition of task complexity, it is possible to categorize noise into two distinct categories: positive noise (PN) and harmful noise (HN). PN decreases the complexity of the task, while HN increases it, aligning with the conventional understanding of noise.

\iffalse
\begin{figure*}[!tp]
\begin{center}
\includegraphics[width=1.0\textwidth]{./figs/BG.pdf}
\end{center}
\caption{The influence of noise on the image. From left to right are the original image, the image with Gaussian noise, overlapping with its own linear transform, and with salt-and-pepper noise, separately.}
\label{noiseBG}
\end{figure*}
\fi

\subsection{ Scope and Contribution }
Our work aims to investigate how various types of noise affect deep learning models. Specifically, the study focuses on three common types of noise, i.e., Gaussian noise, linear transform noise, and salt-and-pepper noise. Gaussian noise refers to random fluctuations that follow a Gaussian distribution in pixel values at the image level or latent representations in latent space \cite{Russo03}. Linear transforms, on the other hand, refer to affine elementary transformations to the dataset of original images or latent representations, where the transformation matrix is row equivalent to an identity matrix \cite{Sun2022CVPR}. Salt-and-pepper noise is a kind of image distortion that adds random black or white values at the image level or to the latent representations \cite{Chan05}.

This paper analyzes the impact of these types of noise on the performance of deep learning models for image classification and domain adaptation tasks. Two popular model families, Vision Transformers (ViTs) and Convolutional Neural Networks (CNNs), are considered in the study. Image classification is one of the most fundamental tasks in computer vision, where the goal is to predict the class label of an input image. Domain adaptation is a practically meaningful task where the training and test data come from different distributions, also known as different domains. By investigating the effects of different types of noise on ViTs and CNNs for typical deep learning tasks, the paper provides insights into the influences of noises on deep models. The findings presented in this paper hold practical significance for enhancing the performance of various types of deep learning models in real-world scenarios.

 The contributions of this paper are summarized as follows: 
\begin{itemize}
    \item We re-examined the conventional view that noise, by default, has a negative impact on deep learning models. Our theoretical analysis and experimental results show that noise can be a positive support for deep learning models and tasks.
    %\item We give out the closed form for the optimal positive noise %for deep models, i.e., the lower boundary and upper boundary of %the benefit from the positive noise to the deep models.
    \item We implemented extensive experiments with different deep models, such as CNNs and ViTs, and on different deep learning tasks. Empowered by positive noise, we achieved state-of-the-art (SOTA) results in the experiments presented in this paper. 
    \item Instead of operating on the image level, our injecting noise operations are performed in the latent space. We theoretically analyze the difference between injecting noise on the image level and in the latent space.
    \item The theory and framework of reducing task complexity via positive noise in this work can be applied to any deep learning architecture. There is great potential for exploring the application of positive noise in other deep-learning tasks beyond the image classification and domain adaptation tasks examined in this study.
\end{itemize}

\subsection{Related Work}
\textbf{Positive Noise} In fact, within the signal-processing society, it has been demonstrated that random noise helps stochastic resonance improve the detection of weak signals \cite{Benzi1981}. Noises can have positive support and contribute to less mean square error compared to the best linear unbiased estimator when the mixing probability distribution is not in the extreme region \cite{Radnosrati2020}. Also, it has been reported that noise could increase the model generalization in natural language processing (NLP) \cite{Pereira2021}. Recently, the perturbation, a special case of positive noise, has been effectively utilized to implement self-refinement in domain adaptation and achieved state-of-the-art performance \cite{Sun2022CVPR}. The latest research shows that by proactively adding specific noise to partial datasets, various tasks can benefit from the positive noise \cite{Li2022Positive}. Besides, noises are found to be able to boost brain power and be useful in many neuroscience studies \cite{McClintock2002} \cite{Mori2002}.

\textbf{Deep Model}
Convolutional Neural Networks have been widely used for image classification, object detection, and segmentation tasks, and have achieved impressive results \cite{LeCun95}\cite{Kaiming16}. However, these networks have limitations in terms of their ability to capture long-range dependencies and extract global features from images. Recently, Vision Transformer has been proposed as an alternative to CNNs \cite{Dosovitskiy20}. ViT relies on self-attention mechanisms and a transformer-based architecture to enable global feature extraction and modeling of long-range dependencies in images \cite{Vaswani17}. The attention mechanism allows the model to focus on the most informative features of the input image, while the transformer architecture facilitates information exchange between different parts of the image. ViT has demonstrated impressive performance on a range of image classification tasks and has the potential to outperform traditional CNN-based approaches. However, ViT currently requires a large number of parameters and training data to achieve state-of-the-art results, making it challenging to implement in certain settings \cite{Zhai22}. 

\section{Preliminary}

In information theory, the entropy \cite{Shannon01} of a random variable $x$ is defined as:
\begin{equation}\label{entropy}
\begin{aligned}
H(x) =\left\{\begin{matrix}
-\int p(x) \log p(x)dx   &  \quad\  \mathrm{if} \ x \ \mathrm{is \ continuous}   \\ 
- \sum_{x} p(x) \log p(x)&   \mathrm{if} \ x \ \mathrm{is \ discrete}
\end{matrix}\right.
\end{aligned}
\end{equation}
where $p(x)$ is the distribution of the given variable $x$. The mutual information of two random discrete variables $(x,y)$ is denoted as \cite{Cover99}:
\begin{equation}\label{MI}
\begin{split}
MI(x,y)=  & D_{KL}(p(x,y)\parallel p(x) \otimes p(y) )  \\
= & H(x) -H(x|y)
\end{split}
\end{equation}
where $D_{KL}$ is the Kullback–Leibler divergence \cite{Kullback51}, and $p(x,y)$ is the joint distribution. The conditional entropy is defined as:
\begin{equation}\label{ce}
H(x|y)=-\sum p(x,y)\log{p(x|y)} 
\end{equation}
The above definitions can be readily expanded to encompass continuous variables through the substitution of the sum operator with the integral symbol. In this work, the noise is denoted by $\boldsymbol{\epsilon }$ if without any specific statement.

Before delving into the correlation between task and noise, it is imperative to address the initial crucial query of the mathematical measurement of a task $\mathcal{T}$. With the assistance of information theory, the complexity associated with a given task $\mathcal{T}$ can be measured in terms of the entropy of $\mathcal{T}$. Therefore, we can borrow the concepts of information entropy to explain the difficulty of the task. For example, a smaller $H({\mathcal{T}})$ means an easier task and vice versa.

Since the entropy of task $\mathcal{T}$ is formulated, it is not difficult to define the entropy change when additional noise $\boldsymbol{\epsilon }$ is present \cite{Li2022Positive},
\begin{equation}\label{MITaskNoise}
\bigtriangleup S(\mathcal{T},\boldsymbol{\epsilon } ) = H(\mathcal{T})-H(\mathcal{T}|\boldsymbol{\epsilon })
\end{equation}
Formally, if the noise can help reduce the complexity of the task, i.e., $H(\mathcal{T}|\boldsymbol{\epsilon }) < H(\mathcal{T})$, then the noise has positive support. Therefore, a noise $\boldsymbol{\epsilon }$ is defined as \textbf{positive noise} (PN) when the noise satisfies $\bigtriangleup S(\mathcal{T},\boldsymbol{\epsilon } ) > 0 $.
On the contrary, when $\bigtriangleup S(\mathcal{T},\boldsymbol{\epsilon } ) \leq 0$, the noise is considered as the conventional noise and named \textbf{harmful noise} (HN).
%The positive noise can be perceived as an augmentation of information gain brought by $\boldsymbol{\epsilon }$.
\begin{equation}\label{positiveNoise}
\left\{\begin{matrix}
\bigtriangleup S(\mathcal{T},\boldsymbol{\epsilon } ) > 0  &  \boldsymbol{\epsilon } \  \mathrm{ is \ positive}  \ \mathrm{noise} \\
\bigtriangleup S(\mathcal{T},\boldsymbol{\epsilon } ) \leq 0  & \boldsymbol{\epsilon } \ \mathrm{is \ harmful}  \ \mathrm{noise}
\end{matrix}\right.
\end{equation}

\iffalse
\textbf{Moderate Noise Assumption}: The existence of positive noise does not indicate that there exists a random variable so that the task can be persistently enhanced with the increase of the positive noise. Even for the positive noise, the conventional consensus of noise does not change: superfluous positive noise will cause degeneration. This is the reason why positive noise is still named "noise".
\fi

\textbf{Moderate Model Assumption}: The positive noise may not work for deep models with severe problems. For example, the model is severely overfitting where models begin to memorize the random fluctuations in the data instead of learning the underlying patterns. In that case, the presence of positive noise will not have a positive impact on improving the models' performance. 

\begin{figure*}[!htp]
\begin{center}
\includegraphics[width=1.0\textwidth]{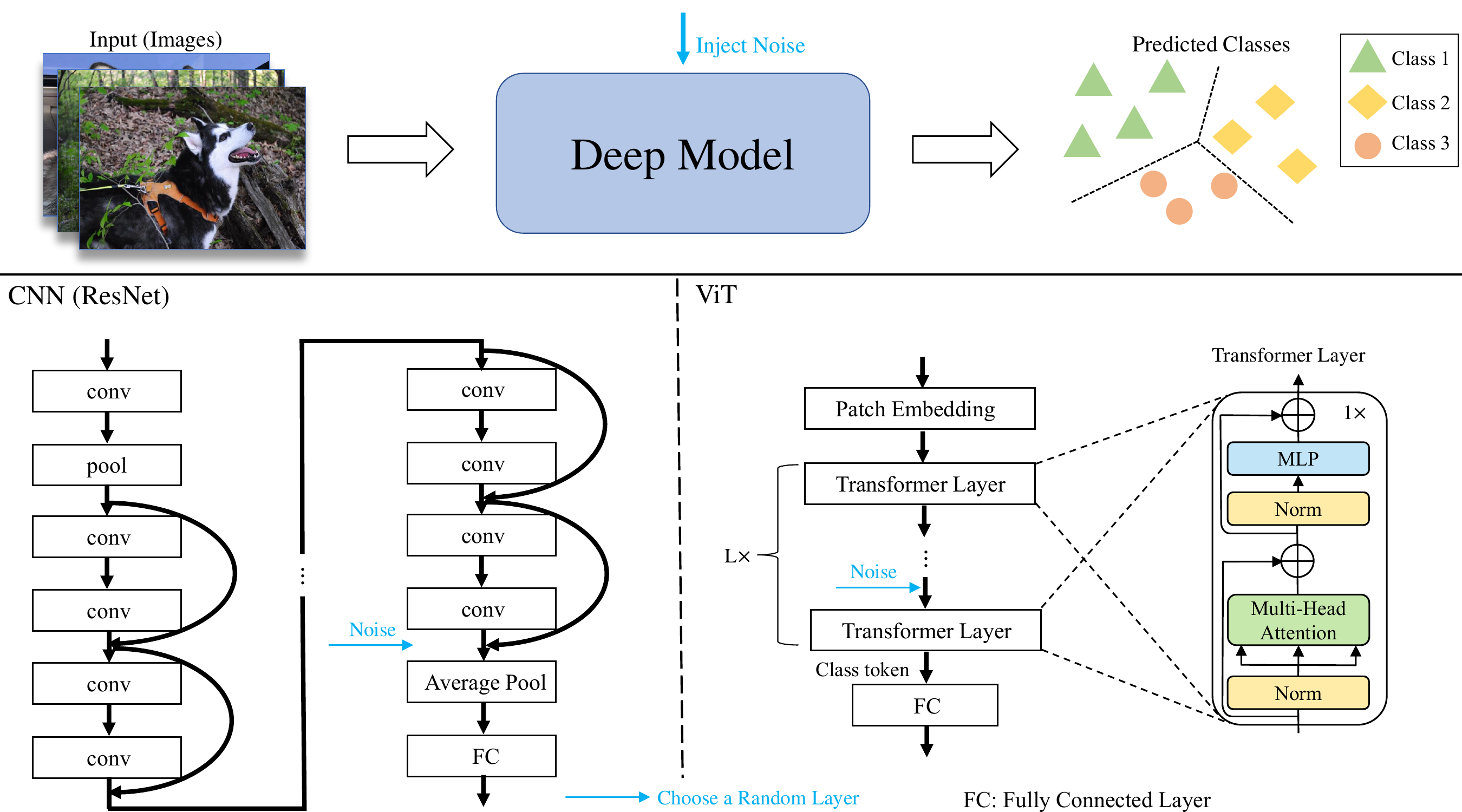}
\end{center}
\caption{An overview of the proposed method. Above the black line is the standard pipeline for image classification. The deep model can be CNNs or ViTs. The noise is injected into a randomly chosen layer of the model represented by the blue arrow. } 
\label{NoisyModel}
\end{figure*}

\section{Methods}
The idea of exploring the influence of noise on the deep models is straightforward. The framework is depicted in Fig. \ref{NoisyModel}. This is a universal framework where there are different options for deep models, such as CNNs and ViTs. Through the simple operation of injecting noise into a randomly selected layer, a model has the potential to gain additional information to reduce task complexity, thereby improving its performance. It is sufficient to inject noise into a single layer instead of multiple layers since it imposes a regularization on multiple layers simultaneously.

For a classification problem, the dataset $(\boldsymbol{X},\boldsymbol{Y})$ can be regarded as samplings derived from $\mathit{D} _{\mathcal{X},\mathcal{Y} } $, where $\mathit{D} _{\mathcal{X},\mathcal{Y} } $ is some unknown joint distribution of data points and labels from feasible space $\mathcal{X}$ and $\mathcal{Y}$, i.e., $(\boldsymbol{X}, \boldsymbol{Y}) \sim \mathit{D} _{\mathcal{X},\mathcal{Y} } $ \cite{Shalev14book}. Hence, given a set of $k$ data points $\boldsymbol{X}=\left \{ X_{1},X_{2},...,X_{k} \right \}$, the label set $\boldsymbol{Y}=\left \{ Y_{1},Y_{2},...,Y_{k} \right \} $ is regarded as sampling from $\boldsymbol{Y} \sim \mathit{D} _{\mathcal{Y}|\mathcal{X} } $. The complexity of $\mathcal{T}$ on dataset $\boldsymbol{X}$ is formulated as:
\begin{equation}\label{complexity}
%H(\mathcal{T};\boldsymbol{X})=H(\boldsymbol{Y};\boldsymbol{X}) - H(\boldsymbol{X})
H_{\mathcal{T}}(\boldsymbol{X}) \coloneqq H(\boldsymbol{Y} \mid \boldsymbol{X})
\end{equation}

Accordingly, the operation of adding noise at the image level can be formulated as \cite{Li2022Positive}:
\begin{align} \label{imglevelcomplexity}
\left\{\begin{matrix}
  H_{\mathcal{T}}(\boldsymbol{X}+\boldsymbol{\epsilon})=-\sum_{\boldsymbol{Y} \in \mathcal{Y}}p(\boldsymbol{Y}|\boldsymbol{X}+\boldsymbol{\epsilon})\log{p(\boldsymbol{Y}|\boldsymbol{X}+\boldsymbol{\epsilon})}   & \boldsymbol{\epsilon} \ \mathrm{is \ additive \ noise}    \\
 H_{\mathcal{T}}(\boldsymbol{X}\boldsymbol{\epsilon})=-\sum_{\boldsymbol{Y} \in \mathcal{Y}}p(\boldsymbol{Y}|\boldsymbol{X}\boldsymbol{\epsilon})\log{p(\boldsymbol{Y}|\boldsymbol{X}\boldsymbol{\epsilon})}    & \boldsymbol{\epsilon} \ \mathrm{is \ multiplicative \ noise} 
\end{matrix}\right.
\end{align}

Inspired by the previous work \cite{Li2022Positive}, given a set of $k$ image embeddings $\boldsymbol{Z}=\left \{ Z_{1},Z_{2},...,Z_{k} \right \}$, the label set $\boldsymbol{Y}=\left \{ Y_{1},Y_{2},...,Y_{k} \right \} $ is regarded as sampling from $\boldsymbol{Y} \sim \mathit{D} _{\mathcal{Y}|\mathcal{Z} } $. The complexity of $\mathcal{T}$ on dataset $\boldsymbol{Z}$ is formulated as:
\begin{equation}\label{latentcomplexity}
%H(\mathcal{T};\boldsymbol{Z}) \coloneqq H(\boldsymbol{Y};\boldsymbol{Z}) - H(\boldsymbol{Z})
H_{\mathcal{T}}(\boldsymbol{Z}) \coloneqq H(\boldsymbol{Y},\boldsymbol{Z}) - H(\boldsymbol{Z})
\end{equation}
We emphasize that $H(\boldsymbol{Z})$ refers to 
the joint entropy of all embeddings in the dataset. Our formulation captures the dataset-level information content rather than per-sample entropy. In this work, we freeze the vision encoder, so the marginal embedding entropy $H(\boldsymbol{Z})$ is kept as a fixed baseline. We then define the task complexity under proactive noise injection in the latent space as:
\begin{align} \label{latentcomplexitynoise}
\left\{\begin{matrix}
  H_{\mathcal{T}}(\boldsymbol{Z}+\boldsymbol{\epsilon})\coloneqq H(\boldsymbol{Y},\boldsymbol{Z}+\boldsymbol{\epsilon}) - H(\boldsymbol{Z})  & \boldsymbol{\epsilon} \ \mathrm{is \ additive \ noise}    \\
 H_{\mathcal{T}}(\boldsymbol{Z}\boldsymbol{\epsilon})\coloneqq H(\boldsymbol{Y},\boldsymbol{Z}\boldsymbol{\epsilon}) - H(\boldsymbol{Z})   & \boldsymbol{\epsilon} \ \mathrm{is \ multiplicative \ noise} 
\end{matrix}\right.
\end{align}
where $\boldsymbol{Z}$ are the embeddings of the images. The definition of Eq. \ref{latentcomplexitynoise} differs from the conventional definition, as our method injects the noise into the latent representations instead of the original images. The Gaussian noise is additive, the linear transform noise is also additive, while the salt-and-pepper noise is multiplicative.

\textbf{Gaussian Noise} The Gaussian noise is one of the most common additive noises that appear in computer vision tasks. The Gaussian noise is independent and stochastic, obeying the Gaussian distribution. Without loss of generality, defined as $\mathcal{N}(\mu, \sigma^{2}  )$. Since our injection happens in the latent space, therefore, the complexity of the task is:
%Gaussian noisy image at the image level is shown in the second picture from left to right in Fig. \ref{noiseBG}. 
\begin{equation}
\begin{matrix}
H_{\mathcal{T}}(\boldsymbol{Z}+\boldsymbol{\epsilon}) = H(\boldsymbol{Y} , \boldsymbol{Z} +\boldsymbol{\epsilon}) -H( \boldsymbol{Z}) %H(\boldsymbol{Y};\boldsymbol{Z}+\boldsymbol{\epsilon}) - H(\boldsymbol{Z}) 
\end{matrix}.
\end{equation}
We assume that both $\boldsymbol{Z}$ and $\boldsymbol{Y}$ follow a multivariate normal distribution. Additionally, we can transform the distributions of $\boldsymbol{Z}$ and $\boldsymbol{Y}$ to make them (approximately) follow the multivariate normal distribution, even if they initially do not \cite{Box64} \cite{Feng2014}. According to the definition in Equation \ref{MITaskNoise}, the entropy change with Gaussian noise is:
\begin{equation} \label{MI_Gaussian}
  \begin{split}
 \bigtriangleup S(\mathcal{T},\boldsymbol{\epsilon } )
  =& H_{\mathcal{T}}(\boldsymbol{Z}) - H_{\mathcal{T}}(\boldsymbol{Z}+\boldsymbol{\epsilon}) \\
 =& H(\boldsymbol{Y} , \boldsymbol{Z} ) - H(\boldsymbol{Z}) - (H(\boldsymbol{Y} , \boldsymbol{Z} +\boldsymbol{\epsilon}) -H(\boldsymbol{Z})) \\
= &  H(\boldsymbol{Y} , \boldsymbol{Z} )  - H(\boldsymbol{Y} , \boldsymbol{Z} +\boldsymbol{\epsilon})  \\
=&\frac{1}{2} \log \frac{|\boldsymbol{\Sigma_{Z}}||\boldsymbol{\Sigma_{Y}-\Sigma_{YZ}{\Sigma}^{-1}_{Z}\Sigma_{ZY}}|}{|\boldsymbol{\Sigma_{Z+\epsilon}}||\boldsymbol{\Sigma_{Y}-\Sigma_{YZ}{\Sigma}^{-1}_{Z+\epsilon}\Sigma_{ZY}}|} \\
= & \frac{1}{2} \log  \frac{ 1}{(1+\sigma^{2}_{\epsilon } {\textstyle \sum_{i=1}^{k}\frac{1}{\sigma^{2}_{Z_{i} }} } )(1+\lambda {\sum_{i=1}^{k} \frac{\mathrm{cov}^{2}(Z_{i},Y_{i})} { \sigma^{2}_{X_{i}}(\sigma^{2}_{Z_{i}}\sigma^{2}_{Y_{i}}-\mathrm{cov}^{2}(Z_{i},Y_{i}))}})}
\end{split} 
\end{equation}
where $\lambda = \frac{\sigma^{2}_{\epsilon }}{1+\sum_{i=1}^{k}\frac{1}{\sigma^{2}_{Z_{i} }}  }$, $\sigma^{2}_{\epsilon }$ is the variance of the Gaussian noise, $\mathrm{cov}(Z_{i},Y_{i})$ is the covariance of sample pair ${X_{i}, Y_{i}}$, $\sigma^{2}_{Z_{i}}$ and $\sigma^{2}_{Y_{i}}$ are the variance of data sample $Z_{i}$ and data label $Y_{i}$, respectively. 

The detailed derivations can be found in section 1.1.2 of the supplementary. Given a dataset, the variance of the Gaussian noise, and statistical properties of data samples and labels control the entropy change, we define the function:
\begin{equation} \label{GauVarianceFunc}
\resizebox{.99\hsize}{!}{$
\begin{split}
M=&1-(1+\sigma^{2}_{\epsilon } {\textstyle \sum_{i=1}^{k}\frac{1}{\sigma^{2}_{Z_{i} }} } )(1+\lambda {\sum_{i=1}^{k} \frac{\mathrm{cov}^{2}(Z_{i},Y_{i})} { \sigma^{2}_{Z_{i}}(\sigma^{2}_{Z_{i}}\sigma^{2}_{Y_{i}}-\mathrm{cov}^{2}(Z_{i},Y_{i}))}}) \\ 
=& - \sigma^{2}_{\epsilon } {\textstyle \sum_{i=1}^{k}\frac{1}{\sigma^{2}_{Z_{i} }} } -\sigma^{2}_{\epsilon } {\textstyle \sum_{i=1}^{k}\frac{1}{\sigma^{2}_{Z_{i} }} } \cdot \lambda{\sum_{i=1}^{k} \frac{\mathrm{cov}^{2}(Z_{i},Y_{i})} { \sigma^{2}_{Z_{i}}(\sigma^{2}_{Z_{i}}\sigma^{2}_{Y_{i}}-\mathrm{cov}^{2}(Z_{i},Y_{i}))}}-\lambda{\sum_{i=1}^{k} \frac{\mathrm{cov}^{2}(Z_{i},Y_{i})} { \sigma^{2}_{Z_{i}}(\sigma^{2}_{Z_{i}}\sigma^{2}_{Y_{i}}-\mathrm{cov}^{2}(Z_{i},Y_{i}))}}
\end{split} 
$}
\end{equation}

Since ${\epsilon }^{2} \ge 0$ and $\lambda \ge 0$, $\sigma^{2}_{Z_{i}}\sigma^{2}_{Y_{i}}-\mathrm{cov}^{2}(Z_{i}, Y_{i})=\sigma^{2}_{Z_{i}}\sigma^{2}_{Y_{i}}(1-\rho_{Z_{i}Y_{i}}^2)\ge 0 $, where $\rho_{Z_{i}Y_{i}}$ is the correlation coefficient, the sign of $M$ is negative. We can conclude that Gaussian noise injected into the latent space is harmful to the task. More details and the Gaussian noise added to the image level are provided in the supplementary.

\textbf{Linear Transform Noise} This type of noise is obtained by elementary transformation of the features matrix, i.e., $\boldsymbol{\epsilon} = Q\boldsymbol{X}$, where $Q$ is a linear transformation matrix. We name the $Q$ the quality matrix since it controls the property of linear transform noise and determines whether positive or harmful. In the linear transform noise injection in the latent space case, the complexity of the task is:
%An example of the influence of linear transform noise on the image is shown in the third picture from left to right in Fig. \ref{noiseBG}.
\begin{equation}\label{LT}
\begin{split}
H_{\mathcal{T}}(\boldsymbol{Z}+Q\boldsymbol{Z})= 
H(\boldsymbol{Y} , \boldsymbol{Z} +Q\boldsymbol{Z}) -H( \boldsymbol{Z} ) 
%H(\boldsymbol{Y};\boldsymbol{Z}+Q\boldsymbol{Z})- H(\boldsymbol{Z} )
\end{split}
\end{equation}
The entropy change is then formulated as:
\begin{equation} \label{LinearNosieFormula}
\begin{split}
 \bigtriangleup S(\mathcal{T},Q\boldsymbol{Z } ) 
 =&H_{\mathcal{T}}(\boldsymbol{Z}) - H_{\mathcal{T}}(\boldsymbol{Z}+Q\boldsymbol{Z})\\
=&H(\boldsymbol{Y} , \boldsymbol{Z} ) - H(\boldsymbol{Y} , \boldsymbol{Z} +Q\boldsymbol{Z})\\
=&\frac{1}{2} \log \frac{|\boldsymbol{\Sigma}_{\boldsymbol{Z}}||\boldsymbol{\Sigma_{Y}}-\boldsymbol{\Sigma_{YZ}}{\boldsymbol{\Sigma_{Z}^{-1}}}\boldsymbol{\Sigma_{ZY}}|}{|\boldsymbol{\Sigma}_{(I+Q)\boldsymbol{Z}}||\boldsymbol{\Sigma_{Y}-\Sigma_{YZ}}\boldsymbol{\Sigma^{-1}_{Z}}\boldsymbol{\Sigma_{XY}}|} \\
=&\frac{1}{2} \log \frac{ 1 }{|I+Q|^{2}}\\
=&-\log|I+Q|
\end{split}
\end{equation}
Since we want the entropy change to be greater than 0, we can formulate Equation \ref{LinearNosieFormula} as an optimization problem:
\begin{equation} \label{LinearNosieOptimization}
\begin{split}
&   \max_{Q}  \bigtriangleup S(\mathcal{T},Q\boldsymbol{Z } ) \\
&s.t. \ rank(I+Q)=k \\
& \ \ \quad Q \sim I \\
&\ \quad  \left [ I+Q \right ]_{ii} \ge  \left [ I+Q \right ]_{ij}, i \ne j \\
& \ \quad  \left \| \left [ I+Q \right ]_{i}   \right \|_{1} = 1 
\end{split}
\end{equation}
where $\sim$ means the row equivalence.
The key to determining whether the linear transform is positive noise or not lies in the matrix of $Q$. The most important step is to ensure that $I+Q$ is reversible, which is $|(I+Q)| \ne 0$. The third constraint is to make the trained classifier get enough information about a specific image and correctly predict the corresponding label. For example, for an image $X_{1}$ perturbed by another image $X_{2}$, the classifier obtained dominant information from $X_{1}$ so that it can predict the label $Y_{1}$. However, if the perturbed image $X_{2}$ is dominant, the classifier can hardly predict the correct label $Y_{1}$ and is more likely to predict as $Y_{2}$. The fourth constraint is to maintain the norm of latent representations. More in-depth discussion and linear transform noise added to the image level are provided in the supplementary. 

\textbf{Salt-and-pepper Noise} The salt-and-pepper noise is a common multiplicative noise for images. The image can exhibit unnatural changes, such as black pixels in bright areas or white pixels in dark areas, specifically as a result of the signal disruption caused by sudden strong interference or bit transmission errors. In the Salt-and-pepper noise case, the entropy change is:
\begin{align*}
\bigtriangleup S(\mathcal{T},\boldsymbol{\epsilon } )
= & H_{\mathcal{T}}(\boldsymbol{Z}) - H_{\mathcal{T}}(\boldsymbol{Z}\boldsymbol{\epsilon})  \\
= & H(\boldsymbol{Y},\boldsymbol{Z}) - H(\boldsymbol{Y},\boldsymbol{Z\epsilon}) \\
= & \mathbb{E}\!\left[\log\frac{1}{p(\boldsymbol{Z},\boldsymbol{Y})}\right] - \mathbb{E}\!\left[\log\frac{1}{p(\boldsymbol{Z\epsilon},\boldsymbol{Y})}\right] \\
\le & \mathbb{E}\!\left[\log\frac{1}{p(\boldsymbol{Z},\boldsymbol{Y})}\right] - \mathbb{E}\!\left[\log\frac{1}{p(\boldsymbol{Z},\boldsymbol{Y})}\right] \;=\; 0
\end{align*}
Obviously, the entropy change is smaller than 0, which indicates the complexity is increasing when injecting salt-and-pepper noise into the deep model. As can be foreseen, the salt-and-pepper noise is pure detrimental noise. More details and Salt-and-pepper added to the image level are in the supplementary.

\iffalse
\textbf{Data augmentation} can be regarded as a special case of positive noise that effectively reduces task complexity. Data augmentation involves applying a set of transformations or modifications to the existing dataset to generate new and diverse data samples \cite{Shorten19}. The idea behind data augmentation is to generate more variations of the same data, which helps the model to learn more features and patterns from the data, without overfitting or memorizing the training data. For example, in image classification tasks, data augmentation techniques can include flipping the image horizontally or vertically, rotating the image, zooming in or out, changing the brightness or contrast, or adding noise to the image \cite{Mikołajczyk18}. By using data augmentation, the deep learning model is trained on a larger and more diverse dataset, which can improve its performance and generalization ability \cite{Perez17}. However, there is lacking theoretical guidance for data augmentation. Our work shed light on effective and efficient data augmentation designs.
\fi

\section{Experiments}
\label{headings}
In this section, we conduct extensive experiments to explore the influence of various types of noises on deep learning models. We employ popular deep learning architectures, including both CNNs and ViTs, and show that the two kinds of deep models can benefit from the positive noise. We employ deep learning models of various scales, including ViT-Tiny (ViT-T), ViT-Small (ViT-S), ViT-Base (ViT-B), and ViT-Large (ViT-L) for Vision Transformers (ViTs), and ResNet-18, ResNet-34, ResNet-50, and ResNet-101 for ResNet architecture. The details of deep models are presented in the supplementary. Without specific instructions, the noise is injected at the last layer of the deep models. Note that for ResNet models, the number of macro layers is 4, and for each macro layer, different scale ResNet models have different micro sublayers. For example, for ResNet-18, the number of macro layers is 4, and for each macro layer, the number of micro sublayers is 2. The noise is injected at the last micro sublayer of the last macro layer for ResNet models. More experimental settings for ResNet and ViT are detailed in the supplementary.

\subsection{Noise Setting}
We utilize the standard normal distribution to generate Gaussian noise in our experiments, ensuring that the noise has zero mean and unit variance. Gaussian noise can be expressed as:
\begin{equation}\label{StarndGaussian}
\begin{split}
\epsilon \sim \mathcal{N}(0,1) 
\end{split}
\end{equation}
For linear transform noise, many possible quality $Q$ matrices could satisfy these constraints, forming a design space. Here, we adopt a simple concrete construction of $Q$ that we call a \emph{circular shift} as a working example. In this construction, each original $Z_i$ is perturbed slightly by its immediate next neighbor $Z_{i+1}$.
We can formally express the circular shift noise injection strategy as follows: Let the scalar hyperparameter $\alpha \in [0,1]$ define the perturbation strength. The quality matrix $Q$ is implemented as $Q = \alpha * U - \alpha * I$, where $U_{i,j} = \delta_{i+1,j}$ with $\delta_{i+1,j}$ representing the Kronecker delta indicator \cite{Frankel11Kronecker}, and employing wrap-around (or ``circular'') indexing. The concrete formation of the quality matrix in circular shift form is then formulated as:
% \begin{align}
%  I + Q =&
% \begin{bmatrix} 
% 1-\alpha &\alpha  & 0  &  0 & 0 \\
%  0 & 1-\alpha & \alpha & 0 & 0 \\
%  0 & 0 & 1-\alpha & \ddots & 0\\
%  0 & 0 & 0 & \ddots & \alpha \\
% \alpha  & 0 & 0 & 0 &1-\alpha
% \label{NoisyNN_PN_generation}
% \end{bmatrix}    
% \end{align}
\begin{align}
 Q =&
\begin{bmatrix} 
-\alpha &\alpha  & 0  &  0 & 0 \\
 0 & -\alpha & \alpha & 0 & 0\\
 0 & 0 & -\alpha & \ddots & 0\\
 0 & 0 & 0 & \ddots & \alpha \\
\alpha  & 0 & 0 & 0 &-\alpha
\label{NoisyNN_PN_generation}
\end{bmatrix}    
\end{align}
The parameter $\alpha$ represents the linear transform strength.

For salt-and-pepper noise, we also use the parameter $\alpha$ to control the probability of the emergence of salt-and-pepper noise, which can be formulated as: 
\begin{equation}\label{ImpulseFormulation}
\begin{split}
\begin{cases}
max(\boldsymbol{Z})  & \text{ if } p<\alpha/2  \\
min(\boldsymbol{Z}) & \text{ if } p>1-\alpha /2
\end{cases}
\end{split}
\end{equation}
where $p$ is a probability generated by a random seed, $ \alpha \in [0,1)$, and $Z$ is the latent representation of an image.

\begin{table}[] 
\caption{ \justifying{ResNet with different kinds of noise on ImageNet. Vanilla means the vanilla model without noise. Accuracy is shown in percentage. Gaussian noise used here is subjected to standard normal distribution. Linear transform noise used in this table is designed to be positive noise. The difference is shown in the bracket.}}
\centering
\begin{tabular}{ccccc}
\hline
Model                       & ResNet-18 & ResNet-34 & ResNet-50 & ResNet-101 \\ \hline
Vanilla                     & 63.90 (+0.00)     & 66.80 (+0.00)    & 70.00 (+0.00)   & 70.66 (+0.00)     \\
+ Gaussian Noise         & 62.35 (-1.55)    & 65.40 (-1.40)     & 69.62 (-0.33)    & 70.10  (-0.56)    \\
+ Linear Transform Noise & \textbf{79.62 (+15.72)}     & \textbf{80.05 (+13.25)}    & \textbf{81.32 (+11.32)}    & \textbf{81.91 (+11.25)}   \\
+ Salt-and-pepper Noise  & 55.45 (-8.45)     & 63.36 (-3.44)    & 45.89 (-24.11)    & 52.96 (-17.70)   \\ \hline
\end{tabular}
\label{ResNetImageNet}
\end{table}

\begin{table}[]
\caption{ ViT with different kinds of noise on ImageNet. Vanilla means the vanilla model without injecting noise. Accuracy is shown in percentage. Gaussian noise used here is subjected to standard normal distribution. Linear transform noise used in this table is designed to be positive noise. The difference is shown in the bracket. Note \textbf{ViT-L is overfitting on ImageNet} \cite{Dosovitskiy20} \cite{Steiner2021}.} 
\centering
\begin{tabular}{ccccc}
\hline
Model                       & ViT-T & ViT-S & ViT-B & ViT-L \\ \hline
Vanilla                     & 79.34 (+0.00)     & 81.88 (+0.00)    & 84.33  (+0.00)   & 88.64 (+0.00)     \\
+ Gaussian Noise         & 79.10 (-0.24)    & 81.80 (-0.08)     & 83.41 (-0.92)    & 85.92  (-2.72)    \\
+ Linear Transform Noise & \textbf{80.69 (+1.35)}     & \textbf{87.27 (+5.39)}    & \textbf{89.99 (+5.66)}    & \textbf{88.72   (+0.08)}   \\
+ Salt-and-pepper Noise  & 78.64 (-0.70)     & 81.75 (-0.13)    & 82.40 (-1.93)    & 85.15  (-3.49)   \\ \hline
\end{tabular}
\label{ViTImageNet}
\end{table}

\begin{table}[]
\caption{ Comparison between Positive Noise Empowered ViT with other ViT variants. Top 1 Accuracy is shown in percentage. Here PN is the positive noise, i.e., linear transform noise.}
\centering
\begin{tabular}{ccccc}
\hline
Model            & Top1 Acc. & Params. & Image Res.                   & Pretrained Dataset \\ \hline
ViT-B \cite{Dosovitskiy20}       & 84.33     & 86M     & 224 $\times$ 224 & ImageNet 21k       \\
DeiT-B \cite{Touvron2021DeiT}        & 85.70     & 86M     & 224 $\times$ 224 & ImageNet 21k       \\
SwinTransformer-B \cite{liu2021Swin} & 86.40     & 88M     & 384 $\times$ 384 & ImageNet 21k       \\
DaViT-B \cite{DingDavit22}          & 86.90     & 88M     & 384 $\times$ 384 & ImageNet 21k       \\
MaxViT-B \cite{TuMaxvit22}      & 88.82     & 119M    & 512 $\times$ 512  & JFT-300M (Private) \\ 
ViT-22B \cite{Dehghani2023}        & 89.51     & 21743M    & 224 $\times$ 224  & JFT-4B (Private) \\ \hline
NoisyViT-B (ViT+B+PN)         & \textbf{89.99}     & 86M     & 224 $\times$ 224 & ImageNet 21k       \\ 
NoisyViT-B (ViT+B+PN)        & \textbf{91.37}     & 86M     & 384 $\times$ 384 & ImageNet 21k       \\ \hline

\end{tabular}\label{ViTcompare}
\end{table}

\begin{table}[]
\caption{ Top 1 accuracy on ImageNet V2 with positive linear transform noise. }
\centering
\begin{tabular}{ccccc}
\hline
Model            & Top1 Acc. & Params. & Image Res.                   & Pretrained Dataset \\ \hline
NoisyViT-B       & \textbf{82.20}     & 86M     & 224 $\times$ 224 & ImageNet 21k       \\ \hline
NoisyViT-B        & \textbf{84.80}     & 86M     & 384 $\times$ 384 & ImageNet 21k       \\ \hline
\end{tabular}\label{PNImageNetV2}
\end{table}

\subsection{Image Classification Results}
We implement extensive experiments on large-scale datasets such as ImageNet \cite{Deng2009} and small-scale datasets such as TinyImageNet \cite{Le2015} using ResNets and ViTs. 

\subsubsection{CNN Family}
The results of ResNets with different noises on ImageNet are in Table \ref{ResNetImageNet}. As shown in the table, with the design of linear transform noise to be positive noise (PN), ResNet improves the classification accuracy by a large margin. While the salt-and-pepper, which is theoretically harmful noise (HN), degrades the models. Note we did not utilize data augmentation techniques for ResNet experiments except for normalization. The significant results show that positive noise can effectively improve classification accuracy by reducing task complexity.

\subsubsection{ViT Family}
The results of ViT with different noises on ImageNet are in Table \ref{ViTImageNet}. Since the ViT-L is overfitting on the ImageNet \cite{Dosovitskiy20} \cite{Steiner2021}, the positive noise did not work well on the ViT-L. As shown in the table, the existence of positive noise improves the classification accuracy of ViT by a large margin compared to vanilla ViT. The comparisons with previously published works, such as DeiT \cite{Touvron2021DeiT}, SwinTransformer \cite{liu2021Swin}, DaViT \cite{DingDavit22}, and MaxViT \cite{TuMaxvit22}, are shown in Table \ref{ViTcompare}, and our positive noise-empowered ViT achieved the new state-of-the-art result. Note that the JFT-300M and JFT-4B datasets are private and not publicly available \cite{Sun17JFT300}, and we believe that ViT large and above will benefit from positive noise significantly if trained on larger JFT-300M or JFT-4B, which is theoretically supported in section 4.4. 

\subsection{Ablation Study}
\begin{figure*}[!htp]
\begin{center}
\includegraphics[width=1.0\textwidth]{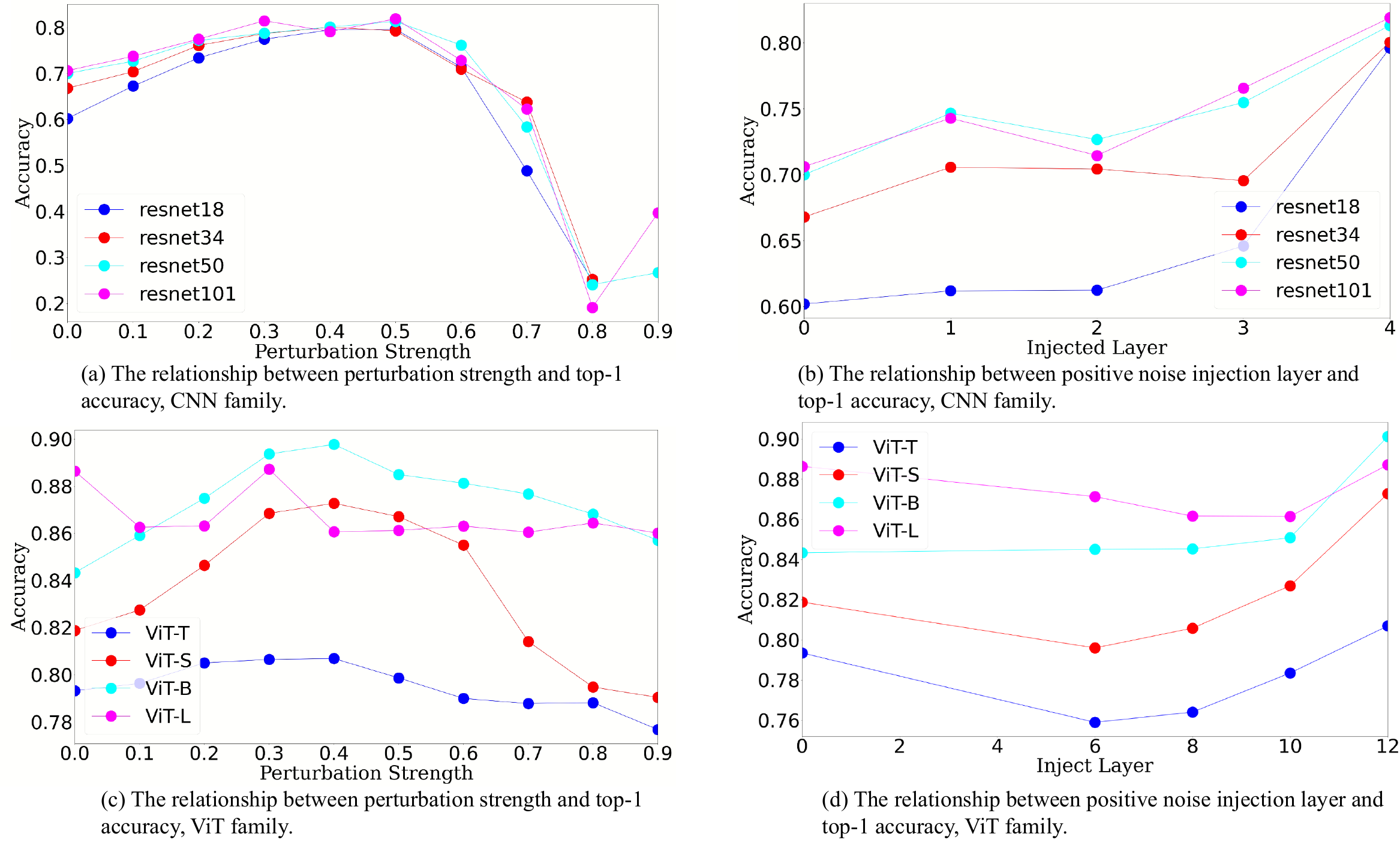}
\end{center}
\caption{The relationship between the linear transform noise strength and the top 1 accuracy, and between the injected layer and top 1 accuracy. Parts (a) and (b) are the results of the CNN family, while parts (c) and (d) are the results of the ViT family. For parts (a) and (c) the linear transform noise is injected at the last layer. For parts (b) and (d), the influence of positive noise on different layers is shown. Layers 6, 8, 10, and 12 in the ViT family are chosen for the ablation study.} 
\label{model}
\label{LinearNoiseStrLayer}
\end{figure*}

We also proactively inject noise into variants of ViT, such as DeiT \cite{Touvron2021DeiT}, Swin Transformer \cite{liu2021Swin}, BEiT \cite{beit}, and ConViT \cite{d2021convit}, and the results show that positive noise could benefit various variants of ViT by improving classification accuracy significantly. The results of injecting noise to variants of ViT are reported in the supplementary. We also did ablation studies on the strength of linear transform noise and the injected layer. The results are shown in Fig. \ref{LinearNoiseStrLayer}. We can observe that the deeper layer the positive noise injects, the better prediction performance the model can obtain. There are reasons behind this phenomenon. First, the latent features of input in the deeper layer have better representations than those in shallow layers; second, injection to shallow layers reduces less task complexity because of trendy replacing Equation \ref{latentcomplexity} with Equation \ref{imglevelcomplexity}. More results on the small dataset TinyImageNet can be found in the supplementary.

Additionally, we tested the positive linear transformation noise on another popular dataset, the ImageNet V2. The corresponding results are reported in Table \ref{PNImageNetV2}.

\begin{table}[]
\caption{ Top 1 accuracy on ImageNet with the optimal quality matrix of linear transform noise. }
\centering
\begin{tabular}{ccccc}
\hline
Model            & Top1 Acc. & Params. & Image Res.                   & Pretrained Dataset \\ \hline
NoisyViT-B+Optimal Q         & \textbf{93.87}     & 86M     & 224 $\times$ 224 & ImageNet 21k       \\ \hline
NoisyViT-B+Optimal Q         & \textbf{95.65}     & 86M     & 384 $\times$ 384 & ImageNet 21k       \\ \hline
\end{tabular}\label{OptiamlQViTB}
\end{table}

\begin{table*}[t]
\caption{ Comparison with various ViT-based methods on \textbf{Office-Home}.} 
\centering
\setlength{\tabcolsep}{0.1mm}{} 
\begin{tabular}{ cccccccccccccc }
\hline
Method        & Ar2Cl & Ar2Pr & Ar2Re & Cl2Ar & Cl2Pr & Cl2Re & Pr2Ar & Pr2Cl & Pr2Re & Re2Ar & Re2Cl & Re2Pr & Avg. \\ \hline
ViT-B \cite{Dosovitskiy20}         & 54.7                & 83.0                & 87.2                & 77.3                & 83.4                & 85.6                & 74.4                & 50.9                & 87.2                & 79.6                & 54.8                & 88.8                & 75.5 \\
TVT-B \cite{Yang23}         & 74.9                & 86.8                & 89.5                & 82.8                & 88.0                & 88.3                & 79.8                & 71.9                & 90.1                & 85.5                & 74.6                & 90.6                & 83.6 \\ 
CDTrans-B \cite{Xu22}     & 68.8                & 85.0                & 86.9                & 81.5                & 87.1                & 87.3                & 79.6                & 63.3                & 88.2                & 82.0                & 66.0                & 90.6                & 80.5 \\ 
SSRT-B \cite{Sun2022CVPR}       & 75.2                & 89.0                & 91.1                & 85.1                & 88.3                & 90.0                & 85.0                & 74.2                & 91.3                & 85.7                & 78.6                & 91.8                & 85.4 \\
NoisyTVT-B &  {\textbf{78.3}}   &  {\textbf{90.6}}     & {\textbf{91.9}}      &  {\textbf{87.8}}       &  {\textbf{92.1}}                & {\textbf{91.9}}     & {\textbf{85.8}}    & {\textbf{78.7}}             & {\textbf{93.0}}   &  {\textbf{88.6}}   &  {\textbf{80.6}}                &  {\textbf{93.5}}    & {\textbf{87.7}} \\ \hline
\end{tabular}
\label{Office-Home}
\end{table*}

\begin{table*}[h]
\caption{ Comparison with various ViT-based methods on \textbf{Visda2017}. }
\centering
\setlength{\tabcolsep}{0.7mm}{}
\begin{tabular}{ cccccccccccccc }
\hline
Method     & plane & bcycl & bus & car & horse & knife & mcycl & person & plant & sktbrd & train & truck & Avg. \\ \hline
ViT-B \cite{Dosovitskiy20}         & 97.7              & 48.1                & 86.6              & 61.6               & 78.1              &  63.4               & 94.7                & 10.3               & 87.7            &   47.7                & 94.4                &  35.5              & 67.1 \\
TVT-B \cite{Yang23}         &   92.9           & 85.6               &77.5               & 60.5              &  93.6             & 98.2            &89.4                & 76.4               & 93.6               & 92.0                & 91.7                & 55.7               & 83.9 \\ 
CDTrans-B \cite{Xu22}     & 97.1                & 90.5               &82.4              & 77.5               & 96.6             & 96.1               &  93.6                &{\textbf{88.6}}               &  {\textbf{97.9}}                & 86.9               &  90.3             & 62.8            & 88.4 \\ 
SSRT-B \cite{Sun2022CVPR}       & {\textbf{98.9}}               & 87.6              & { \textbf{89.1} }         &{ \textbf{84.8} }     & 98.3      & {\textbf{98.7} }      &{\textbf{96.3} }               & 81.1               &94.9             & 97.9        & 94.5                & 43.1               & 88.8 \\
NoisyTVT-B &  98.8   &  {\textbf{95.5}}     &84.8    &  73.7     &  {\textbf{98.5}}                & 97.2     & 95.1    & 76.5    & 95.9   &  {\textbf{98.4}}   &  {\textbf{98.3}}                &  {\textbf{67.2}}    & {\textbf{90.0}} \\ \hline
\end{tabular}
\label{Visda17}
\end{table*}

\subsection{Optimal Quality Matrix}
As shown in Equation \ref{LinearNosieOptimization}, it is interesting to learn about the optimal quality matrix of $Q$ that maximizes the entropy change while satisfying the constraints. This equals minimizing the determinant of the matrix sum of $I$ and $Q$. Here, we directly give out the optimal quality matrix of $Q$ as:
\begin{equation}\label{OptimalQ}
Q_{optimal}=\mathrm{diag} \left ( \frac{1}{k+1}-1, \dots, \frac{1}{k+1}-1  \right ) +\frac{1}{k+1} \boldsymbol{1}_{k\times k}
\end{equation}
where $k$ is the number of data samples. And the corresponding upper boundary of the entropy change as:
\begin{equation}\label{UpperBoundaryQ}
 \bigtriangleup S(\mathcal{T},Q_{optim al}\boldsymbol{X } ) = (k-1)\log{(k+1) }
\end{equation}
The details are provided in the supplementary. We find that the upper boundary of the entropy change of injecting positive noise is determined by the number of data samples, i.e., the scale of the dataset. Therefore, the larger the dataset, the better effect of injecting positive noise into deep models. With the optimal quality matrix and the top 1 accuracy of ViT-B on ImageNet can be further improved to 95$\%$, which is shown in Table \ref{OptiamlQViTB}.

\subsection{Domain Adaption Results}
Unsupervised domain adaptation (UDA) aims to learn transferable knowledge across the source and target domains with different distributions \cite{Pan09Survey} \cite{Ying18Transfer}. Recently, transformer-based methods achieved SOTA results on UDA, therefore, we evaluate the ViT-B with the positive noise on widely used UDA benchmarks. Here the positive noise is the linear transform noise identical to that used in the classification task. The positive noise is injected into the last layer of the model, the same as the classification task. The datasets include \textbf{Office Home} \cite{Venkateswara17} and \textbf{VisDA2017} \cite{Peng17}. Detailed datasets introduction and experiments training settings are in the supplementary. The objective function is borrowed from TVT \cite{Yang23}, which is the first work that adopts Transformer-based architecture for UDA. The results are shown in Table \ref{Office-Home} and \ref{Visda17}. The NoisyTVT-B, i.e., TVT-B with positive noise, achieves better performance than existing works. These results demonstrate that positive noise also works in domain adaptation tasks, where out-of-distribution (OOD) data exists.

\section{Conclusion}
This study delves into the influence of entropy change on learning systems, achieved by proactively introducing various types of noise into deep models. Our work conducts a comprehensive investigation into the impact of common noise types, such as Gaussian noise, linear transform noise, and salt-and-pepper noise, on deep learning models. Notably, we demonstrate that, under specific conditions, linear transform noise can positively affect deep models. The experimental results show that injecting positive noise into the latent space significantly enhances the prediction performance of deep models in image classification tasks, leading to new state-of-the-art results on ImageNet. These findings hold broad implications for future research and the potential development of more accurate models for improved real-world applications. 

\section{Potential Impact}
The proposed theory exhibits versatility, opening avenues for exploring the application of positive noise across diverse deep learning tasks in computer vision and natural language processing. While our current findings and theoretical analysis are centered around addressing classification problems in the computer vision domain, their implications extend far beyond. Specifically, the insights hold substantial promise for various learning tasks, including those involving large language models (LLMs). The optimal quality matrix derived from the linear transform noise suggests a potential for more effective model enhancement with larger datasets. This finding implies that the principles uncovered in our study may contribute to refining the strategies for improving large-scale language models. The adaptability of the proposed theory to both computer vision and NLP tasks marks a promising avenue for future research. A potential concern is that those who possess high-quality large-scale datasets may primarily benefit from our research.

\clearpage
\appendix
\begin{appendices}

\section{Theoretical Foundations of Task Entropy}
This section provides the theoretical foundations of task entropy, quantifying the complexity of learning tasks. The concept of task entropy was first proposed for the image level and formulated as \cite{Li2022Positive}:
\begin{equation}
H_{\mathcal{T}}(\boldsymbol{X})=-\sum_{\boldsymbol{Y} \in \mathcal{Y}}p(\boldsymbol{Y}|\boldsymbol{X})\log{p(\boldsymbol{Y}|\boldsymbol{X})}  
\end{equation}

\noindent The image $\boldsymbol{X}$ in the dataset are supposed to be independent of each other, as are the labels $\boldsymbol{Y}$. However, $\boldsymbol{X}$ and $\boldsymbol{Y}$ are not independent because of the correlation between a data sample $X$ and its corresponding label $Y$. Essentially, the task entropy is the entropy of $p(Y|X)$. Following the principle of task entropy, compelling evidence suggests that diminishing task complexity via reducing information entropy can enhance overall model performance \cite{Li2022Positive, Zhang2023Positive}.

Inspired by the concept of task entropy at the image level, we explore its extension to the latent space. The task entropy from the perspective of embeddings is defined as:
\begin{equation}
H_{\mathcal{T}}(\boldsymbol{Z}) \coloneqq H(\boldsymbol{Y},\boldsymbol{Z}) - H(\boldsymbol{Z})
\end{equation}
where $\boldsymbol{Z}$ are the embeddings of the images $\boldsymbol{X}$. Here, we assume that the embedding $\boldsymbol{Z}$ and the vectorized label $\boldsymbol{Y}$ follow a multivariate normal distribution. We can transform the unknown distributions of $\boldsymbol{Z}$ and $\boldsymbol{Y}$ to approximately conform to normality by utilizing techniques if they are not initially normal. After approximate transformation, the distribution of $\boldsymbol{Z}$ and $\boldsymbol{Y}$ can be expressed as:
\begin{equation}\label{SlackXY}
 %\boldsymbol{X}\sim \mathcal{N}(\mu_{X}, \sigma^{2}_{X}  ), %\boldsymbol{Y}\sim \mathcal{N}(\mu_{Y}, \sigma^{2}_{Y}  )
  \boldsymbol{Z}\sim \mathcal{N}( \boldsymbol{\mu_{Z}}, \boldsymbol{\Sigma_{Z}}  ), \boldsymbol{Y}\sim \mathcal{N}(\boldsymbol{\mu_{Y}}, \boldsymbol{\Sigma_{Y}}  )
\end{equation}
where
\begin{equation}
\begin{split}
  &\boldsymbol{{\mu}_{Z} }=\mathbb{E}[\boldsymbol{Z}]=(\mathbb{E}[Z_{1}], \mathbb{E}[Z_{2}],...,\mathbb{E}[Z_{k}]])^{T} \\
  &\boldsymbol{{\mu}_{Y} }=\mathbb{E}[\boldsymbol{Y}]=(\mathbb{E}[Y_{1}], \mathbb{E}[Y_{2}],...,\mathbb{E}[Y_{k}]])^{T} \\
  &\boldsymbol{\Sigma_{Z}}=\mathbb{E}[(\boldsymbol{Z}-\boldsymbol{{\mu}_{Z}})(\boldsymbol{Z}-\boldsymbol{{\mu}_{Z}})^{T}] \\
 & \boldsymbol{\Sigma_{Y}}=\mathbb{E}[(\boldsymbol{Y}-\boldsymbol{{\mu}_{Y}})(\boldsymbol{Y}-\boldsymbol{{\mu}_{Y}})^{T}]
\end{split}
\end{equation}
$k$ is the number of samples in the dataset, and $T$ represents the transpose of the matrix.

Then the conditional distribution of $\boldsymbol{Y}$ given $\boldsymbol{Z}$ is also normally distributed \cite{Mood1950} \cite{Johnson1995}, which can be formulated as:
\begin{equation}
\boldsymbol{Y}|\boldsymbol{Z}\sim \mathcal{N}(\mathbb{E}({\boldsymbol{Y}|\boldsymbol{Z}={Z}}), var(\boldsymbol{Y}|\boldsymbol{Z}={Z}))
\end{equation}
where $\mathbb{E}({\boldsymbol{Y}|\boldsymbol{Z}=Z})$ is the mean of the label set $\boldsymbol{Y}$ given a sample $\boldsymbol{Z}=Z$ from the embeddings, and $var(\boldsymbol{Y}|\boldsymbol{Z}=Z)$ is the variance of $\boldsymbol{Y}$ given a sample from the embeddings.
The conditional mean $\mathbb{E}[(\boldsymbol{Y}|\boldsymbol{Z}=Z) ] $ and conditional variance $var(\boldsymbol{Y}|\boldsymbol{Z}=Z)$ can be calculated as: 
\begin{equation}
\boldsymbol{\mu}_{\boldsymbol{Y}|\boldsymbol{Z}=Z}=\mathbb{E}[(\boldsymbol{Y}|\boldsymbol{Z}=Z) ] =\boldsymbol{{\mu}_{Y}}+\boldsymbol{\Sigma_{YZ}}\boldsymbol{{\Sigma}^{-1}_{Z}}(Z-\boldsymbol{\mu_{Z}})
\end{equation}
\begin{equation}
\boldsymbol{\Sigma}_{{\boldsymbol{Y|Z}}=Z}=var(\boldsymbol{Y}|\boldsymbol{Z}=Z)=\boldsymbol{\Sigma_{Y}-\Sigma_{YX}{\Sigma}^{-1}_{Z}\Sigma_{ZY}}
\end{equation}
where $\boldsymbol{\Sigma_{YZ}}$ and $\boldsymbol{\Sigma_{ZY}}$ are the cross-covariance matrices between $\boldsymbol{Y}$ and $\boldsymbol{Z}$, and between $\boldsymbol{Z}$ and $\boldsymbol{Y}$, respectively, and $\boldsymbol{\Sigma^{-1}_{Z}}$ denotes the inverse of the covariance matrix of $\boldsymbol{Z}$.

\noindent Now, we shall obtain the task entropy:
\begin{equation}
\begin{split}
H_{\mathcal{T}}(\boldsymbol{Z})=&-\sum_{\boldsymbol{Y} \in \mathcal{Y}}p(\boldsymbol{Y}|\boldsymbol{Z})\log{p(\boldsymbol{Y}|\boldsymbol{Z})}  \\
=&-\mathbb{E}[\log p(\boldsymbol{Y}|\boldsymbol{Z})] \\
=& -\mathbb{E}[\log [(2\pi)^{-k/2}|\boldsymbol{\Sigma_{Z}}|^{-1/2}\exp(-\frac{1}{2} \boldsymbol{(Y|Z - \mu_{Y|Z})}^{T} \boldsymbol{\Sigma}^{-1}_{\boldsymbol{Y|Z}}(\boldsymbol{Y|Z} - \boldsymbol{\mu_{Y|Z}}))]] \\
=&\frac{k}{2} \log (2\pi) + \frac{1}{2} \log|\boldsymbol{\Sigma_{Y|Z}}|+\frac{1}{2} \mathbb{E}[{(\boldsymbol{Y|Z} - \boldsymbol{\mu}_{\boldsymbol{Y|Z}})}^{T} \Sigma^{-1}_{\boldsymbol{\boldsymbol{Y|Z}}}(\boldsymbol{Y|Z} - \boldsymbol{\mu}_{\boldsymbol{Y|Z}})] \\
=&\frac{k}{2} (1+ \log (2\pi)) + \frac{1}{2} \log |\boldsymbol{\Sigma_{Y|Z}}|
\end{split}
\end{equation}
where
\begin{equation}
\begin{split}
 \mathbb{E}[{(\boldsymbol{Y|Z} - \boldsymbol{\mu}_{\boldsymbol{Y|Z}})}^{T} \Sigma^{-1}_{\boldsymbol{\boldsymbol{Y|Z}}}(\boldsymbol{Y|Z} - \boldsymbol{\mu}_{\boldsymbol{Y|Z}})]=& \mathbb{E}[tr((\boldsymbol{Y|Z} - \boldsymbol{\mu}_{\boldsymbol{Y|Z}})^{T} \Sigma^{-1}_{\boldsymbol{\boldsymbol{Y|Z}}}(\boldsymbol{Y|Z} - \boldsymbol{\mu}_{\boldsymbol{Y|Z}}))] \\
=& \mathbb{E}[tr (\Sigma^{-1}_{\boldsymbol{\boldsymbol{Y|Z}}} (\boldsymbol{Y|Z} - \boldsymbol{\mu}_{\boldsymbol{Y|Z}}){(\boldsymbol{Y|Z} - \boldsymbol{\mu}_{\boldsymbol{Y|Z}})}^{T})]\\
=& tr (\Sigma^{-1}_{\boldsymbol{\boldsymbol{Y|Z}}} (\boldsymbol{Y|Z} - \boldsymbol{\mu}_{\boldsymbol{Y|Z}}){(\boldsymbol{Y|Z} - \boldsymbol{\mu}_{\boldsymbol{Y|Z}})}^{T}) \\
=&tr (\Sigma^{-1}_{\boldsymbol{\boldsymbol{Y|Z}}} \Sigma_{\boldsymbol{\boldsymbol{Y|Z}}} ) \\
=&tr (\boldsymbol{I}_{k} ) \\
=&k
\end{split}
\end{equation} Therefore, for a specific set of embeddings, we can find that the task entropy is only related to the variance of the $\boldsymbol{Y|Z}$.

As we proactively inject different kinds of noises into the latent space, the task entropy with noise injection is defined as :
\begin{align} \label{latentEntropyChange}
\left\{\begin{matrix}
H_{\mathcal{T}}(\boldsymbol{Z}+\boldsymbol{\epsilon})\coloneqq H(\boldsymbol{Y},\boldsymbol{Z}+\boldsymbol{\epsilon}) - H(\boldsymbol{Z})  & \boldsymbol{\epsilon} \ \mathrm{is \ additive \ noise}    \\
 H_{\mathcal{T}}(\boldsymbol{Z}\boldsymbol{\epsilon})\coloneqq H(\boldsymbol{Y},\boldsymbol{Z}\boldsymbol{\epsilon}) - H(\boldsymbol{Z})   & \boldsymbol{\epsilon} \ \mathrm{is \ multiplicative \ noise} 
\end{matrix}\right.
\end{align}
Equation \ref{latentEntropyChange} diverges from the conventional definition of conditional entropy as our method introduces noise into the latent representations instead of the original images. The noises examined in this study are classified into additive and multiplicative categories. In the subsequent sections, we analyze the changes in task entropy resulting from the injection of common noises into the embeddings.

\iffalse
If adding noise to the original images, then we have the classic definition:
\begin{align}
\left\{\begin{matrix}
H(\mathcal{T};\boldsymbol{X}+\boldsymbol{\epsilon})=H(\boldsymbol{Y};\boldsymbol{X}+\boldsymbol{\epsilon}) - H(\boldsymbol{X}+\boldsymbol{\epsilon}) & \boldsymbol{\epsilon} \ \mathrm{is \ additive \ noise}  \\
H(\mathcal{T};\boldsymbol{X}\boldsymbol{\epsilon})=H(\boldsymbol{Y};\boldsymbol{X}\boldsymbol{\epsilon}) - H(\boldsymbol{X}\boldsymbol{\epsilon})   & \boldsymbol{\epsilon} \ \mathrm{is \ multiplicative \ noise}
\end{matrix}\right.
\end{align}
Examples of the influence of various noises on the image level are provided in Fig. \ref{noiseBG}.

\begin{figure*}[!tp]
\begin{center}
\includegraphics[width=1.0\textwidth]{./figs/BG.pdf}
\end{center}
\caption{The influence of noise on the image. From left to right are the original image, the image with Gaussian noise, overlapping with its own linear transform, and with salt-and-pepper noise, separately.}
\label{noiseBG}
\end{figure*}
\fi

\section{The Impact of Gaussian Noise on Task Entropy}
\label{GaussianCase}
We begin by examining the impact of Gaussian noise on task entropy from the perspective of latent space.

\subsection{Inject Gaussian Noise in Latent Space}
In this case, the task complexity is formulated as:
\begin{equation}
\begin{matrix}
H_{\mathcal{T}}(\boldsymbol{Z}+\boldsymbol{\epsilon})= H(\boldsymbol{Y},\boldsymbol{Z}+\boldsymbol{\epsilon}) - H(\boldsymbol{Z})
\end{matrix}.
\end{equation}
\noindent Take advantage of the definition of task entropy, thus, the entropy change of injecting Gaussian noise in the latent space can be formulated as:
\begin{equation}
  \begin{split}
 \bigtriangleup S(\mathcal{T},\boldsymbol{\epsilon } ) =& H_{\mathcal{T}}(\boldsymbol{Z}) - H_{\mathcal{T}}(\boldsymbol{Z}+\boldsymbol{\epsilon})\\=& H(\boldsymbol{Y}, \boldsymbol{Z}) -H(\boldsymbol{Z}) - (H(\boldsymbol{Y}, \boldsymbol{Z}+\boldsymbol{\epsilon} ) -H(\boldsymbol{Z} ))\\
=& \frac{1}{2} \log|\boldsymbol{\Sigma_{Y|Z}}| + \frac{1}{2} \log|\boldsymbol{\Sigma_{Z}}|- \frac{1}{2} \log|\boldsymbol{\Sigma_{Y|Z+\epsilon}}| -\frac{1}{2} \log|\boldsymbol{\Sigma_{Z+\boldsymbol{\epsilon}}}| \\
=& \frac{1}{2} \log \frac{|\boldsymbol{\Sigma_{Z}}||\boldsymbol{\Sigma_{Y|Z}}|}{|\boldsymbol{\Sigma_{Z+\epsilon}}||\boldsymbol{\Sigma_{Y|Z+\epsilon}}|} \\
=& \frac{1}{2} \log \frac{|\boldsymbol{\Sigma_{Z}}||\boldsymbol{\Sigma_{Y}-\Sigma_{YZ}{\Sigma}^{-1}_{Z}\Sigma_{ZY}}|}{|\boldsymbol{\Sigma_{Z+\epsilon}}||\boldsymbol{\Sigma_{Y}-\Sigma_{YZ}{\Sigma}^{-1}_{Z+\epsilon}\Sigma_{ZY}}|}
\end{split} 
\end{equation}
where $ \boldsymbol{\Sigma_{Y|Z+\epsilon}}=\boldsymbol{\Sigma_{Y}-\Sigma_{Y(Z+\epsilon )}{\Sigma}^{-1}_{Z+\epsilon }\Sigma_{(Z+\epsilon )Y}}$. Since the Gaussian noise is independent of $\boldsymbol{Z}$ and $\boldsymbol{Y}$, we have $\boldsymbol{\Sigma_{Y(Z+\epsilon)}}= \boldsymbol{\Sigma_{(Z+\epsilon)Y}} =\boldsymbol{\Sigma_{YZ}}$. The corresponding proof is:
\begin{equation}
    \begin{split}
\boldsymbol{\Sigma_{(Z+\epsilon)Y}} =& \mathbb{E} \left [   \boldsymbol{(Z+\epsilon)} - \mu_{\boldsymbol{Z+\epsilon}}\right ] \mathbb{E} \left [   \boldsymbol{Y} -\mu_{\boldsymbol{Y}}\right ]\\
=&\mathbb{E}\left [ \boldsymbol{(Z+\epsilon)} \boldsymbol{Y}\right ] -\mu_{\boldsymbol{Y}}\mathbb{E} \left [   \boldsymbol{(Z+\epsilon)} \right ]-\mu_{\boldsymbol{Z+\epsilon}}\mathbb{E} \left [   \boldsymbol{Y}\right ]+\mu_{\boldsymbol{Y}}\mu_{\boldsymbol{Z+\epsilon}}\\
=&\mathbb{E}\left [ \boldsymbol{(Z+\epsilon)} \boldsymbol{Y}\right ]-\mu_{\boldsymbol{Y}}\mathbb{E} \left [   \boldsymbol{(Z+\epsilon)} \right ] \\
=& \mathbb{E}\left [ \boldsymbol{Z} \boldsymbol{Y}\right ]+\mathbb{E}\left [  \boldsymbol{\epsilon}\boldsymbol{Y}\right ]-\mu_{\boldsymbol{Y}}\mu_{\boldsymbol{Z}}- \mu_{\boldsymbol{Y}}\mu_{\boldsymbol{\epsilon }}\\
=& \mathbb{E}\left [ \boldsymbol{Z} \boldsymbol{Y}\right ]- \mu_{\boldsymbol{Y}}\mu_{\boldsymbol{Z}}\\
=&\boldsymbol{\Sigma_{ZY}}
\end{split}
\end{equation}

Obviously, 
\begin{equation} \label{LatentGNJudge}
\left\{\begin{matrix}
\bigtriangleup S(\mathcal{T},\boldsymbol{\epsilon } ) >0 & if \ \frac{|\boldsymbol{\Sigma_{Z}}||\boldsymbol{\Sigma_{Y}-\Sigma_{YZ}{\Sigma}^{-1}_{Z}\Sigma_{ZY}}|}{|\boldsymbol{\Sigma_{Z+\epsilon}}||\boldsymbol{\Sigma_{Y}-\Sigma_{YZ}{\Sigma}^{-1}_{Z+\epsilon}\Sigma_{ZY}}|} > 1\\
\bigtriangleup S(\mathcal{T},\boldsymbol{\epsilon } ) \leq 0  & if \ \frac{|\boldsymbol{\Sigma_{Z}}||\boldsymbol{\Sigma_{Y}-\Sigma_{YZ}{\Sigma}^{-1}_{Z}\Sigma_{ZY}}|}{|\boldsymbol{\Sigma_{Z+\epsilon}}||\boldsymbol{\Sigma_{Y}-\Sigma_{YZ}{\Sigma}^{-1}_{Z+\epsilon}\Sigma_{ZY}}|} \leq 1
\end{matrix}\right.
\end{equation}

\noindent To find the relationship between $|\boldsymbol{\Sigma_{Z}}||\boldsymbol{\Sigma_{Y|Z}}|$ and $|\boldsymbol{\Sigma_{Z+\epsilon}}||\boldsymbol{\Sigma_{Y|Z+\epsilon}}|$, we need to determine the subterms in each of them. As we mentioned in the previous section, the embeddings of the images are independent of each other, and so are the labels.
\begin{equation}
    \begin{split}
 \boldsymbol{\Sigma_{Y}}=&\mathbb{E}[(\boldsymbol{Y}-\boldsymbol{{\mu}_{Y}})(\boldsymbol{Y}-\boldsymbol{{\mu}_{Y}})^{T}]\\
=&\mathbb{E}[\boldsymbol{Y}\boldsymbol{Y}^{T}] -\boldsymbol{{\mu}_{Y}}\boldsymbol{{\mu}_{Y}}^{T} \\
=&\mathrm{diag}( \sigma^{2}_{Y_{1}},...,\sigma^{2}_{Y_{k}})
\end{split}
\end{equation}
where 
\begin{equation}
    \left\{\begin{matrix}
 \mathbb{E}\left [ Y_{i}Y_{j} \right ]-\mu_{Y_{i}} \mu_{Y_{j}}= 0, & i\ne j \\
 \mathbb{E}\left [ Y_{i}Y_{j} \right ]-\mu_{Y_{i}} \mu_{Y_{j}} = \sigma^{2}_{Y_{i}}, & i=j \
\end{matrix}\right.
\end{equation}
The same procedure can be applied to $\boldsymbol{\Sigma_{Y(Z+\epsilon) }}$ and $\boldsymbol{{\Sigma}_{Z+\epsilon}}$. Therefore, We can obtain that $\boldsymbol{\Sigma_{Y}}=\mathrm{diag}( \sigma^{2}_{Y_{1}},...,\sigma^{2}_{Y_{k}})$,
\begin{equation}
    \boldsymbol{\Sigma_{Y(Z+\epsilon) }}=\mathrm{diag} (\mathrm{cov} (Y_{1},Z_{1}+\epsilon ),...,\mathrm{cov} (Y_{k},Z_{k}+\epsilon ))
\end{equation}
and $\boldsymbol{{\Sigma}_{Z+\epsilon}}$ is:
\begin{equation}
\begin{split} \label{epsilonwithXcovariance}
\boldsymbol{{\Sigma}_{Z+\epsilon}}=&\begin{bmatrix}
 \sigma^{2}_{Z_{1}}+\sigma^{2}_{\epsilon } & \sigma^{2}_{\epsilon }  & ...  & \sigma^{2}_{\epsilon } &\sigma^{2}_{\epsilon } \\
\sigma^{2}_{\epsilon } &  \sigma^{2}_{Z_{2}}+\sigma^{2}_{\epsilon }  & ...  & \sigma^{2}_{\epsilon } &\sigma^{2}_{\epsilon }\\
 \vdots  &  \vdots&  &  \vdots & \vdots \\
\sigma^{2}_{\epsilon } &\sigma^{2}_{\epsilon }  & ... &   \sigma^{2}_{Z_{k-1}}+\sigma^{2}_{\epsilon } &\sigma^{2}_{\epsilon }\\
\sigma^{2}_{\epsilon } & \sigma^{2}_{\epsilon }  & ...  & \sigma^{2}_{\epsilon } & \sigma^{2}_{Z_{k}}+\sigma^{2}_{\epsilon }\\
\end{bmatrix} \\
=&\mathrm{diag} ( \sigma^{2}_{Z_{1}},..., \sigma^{2}_{Z_{k}})\boldsymbol{I}_{k}+\sigma^{2}_{\epsilon }\boldsymbol{1}_{k}
\end{split}
\end{equation}
where $\boldsymbol{I}_{k}$ is a $k\times k$ identity matrix and $\boldsymbol{1}_{k}$ is a all ones $k\times k$ matrix. We use $\boldsymbol{U}$ to represent $\mathrm{diag} ( \sigma^{2}_{Z_{1}},..., \sigma^{2}_{Z_{k}})\boldsymbol{I}_{k}$, and $\boldsymbol{u}$ to represent a all ones vector $[1,...,1]^{T}$. Thanks to the Sherman–Morrison Formula \cite{Sherman1949} and Woodbury Formula \cite{Woodbury1950}, we can obtain the inverse of $\boldsymbol{{\Sigma}_{Z+\epsilon}}$ as:
\begin{equation} \label{XplusepsionInverse}
\begin{split}
\boldsymbol{{\Sigma}_{Z+\epsilon}^{-1}}=&(\boldsymbol{U}+\sigma^{2}_{\epsilon }\boldsymbol{ uu }^{T})^{-1} \\
=&\boldsymbol{U}^{-1}-\frac{\sigma^{2}_{\epsilon }}{1+\sigma^{2}_{\epsilon }\boldsymbol{u}^{T}\boldsymbol{U}^{-1}\boldsymbol{u}} \boldsymbol{U}^{-1}\boldsymbol{u}\boldsymbol{u}^{T}\boldsymbol{U}^{-1} \\
=& \boldsymbol{U}^{-1}-\frac{\sigma^{2}_{\epsilon }}{1+\sum_{i=1}^{k}\frac{1}{\sigma^{2}_{Z_{i} }}  } \boldsymbol{U}^{-1}\boldsymbol{1}_{k}\boldsymbol{U}^{-1}  \\
=& \lambda \begin{bmatrix}
\frac{ 1 }{\lambda\sigma^{2}_{Z_{1}}}-\frac{1}{\sigma^{4}_{Z_{1}}}   & -\frac{1}{\sigma^{2}_{Z_{1}}\sigma^{2}_{Z_{2}}}  & ...  & -\frac{1}{\sigma^{2}_{Z_{1}}\sigma^{2}_{Z_{k-1}}} &-\frac{1}{\sigma^{2}_{Z_{1}}\sigma^{2}_{Z_{k}}}\\
-\frac{1}{\sigma^{2}_{Z_{2}}\sigma^{2}_{Z_{1}}}  & \frac{ 1 }{\lambda\sigma^{2}_{Z_{2}}} -\frac{1}{\sigma^{4}_{Z_{2}}}   & ...  & -\frac{1}{\sigma^{2}_{Z_{2}}\sigma^{2}_{Z_{k-1}}}  &-\frac{1}{\sigma^{2}_{Z_{2}}\sigma^{2}_{Z_{k}}} \\
 \vdots  &  \vdots&  &  \vdots & \vdots \\
-\frac{1}{\sigma^{2}_{Z_{k-1}}\sigma^{2}_{Z_{1}}}  &-\frac{1}{\sigma^{2}_{Z_{k-1}}\sigma^{2}_{Z_{2}}}   & ... & \frac{ 1 }{\lambda\sigma^{2}_{Z_{k-1}}}  -\frac{1}{\sigma^{4}_{Z_{k-1}}}  &-\frac{1}{\sigma^{2}_{Z_{k-1}}\sigma^{2}_{Z_{k}}} \\
-\frac{1}{\sigma^{2}_{Z_{k}}\sigma^{2}_{Z_{1}}}  & -\frac{1}{\sigma^{2}_{Z_{k}}\sigma^{2}_{Z_{2}}}   & ...  & -\frac{1}{\sigma^{2}_{Z_{k}}\sigma^{2}_{Z_{k-1}}}  &\frac{ 1 }{\lambda\sigma^{2}_{Z_{k}}} -\frac{1}{\sigma^{4}_{Z_{k}}} \\
\end{bmatrix} 
\end{split}
\end{equation}
where $\boldsymbol{U}^{-1} =\mathrm{diag} ( (\sigma^{2}_{Z_{1}})^{-1},..., (\sigma^{2}_{Z_{k}})^{-1} ) $ and $\lambda = \frac{\sigma^{2}_{\epsilon }}{1+\sum_{i=1}^{k}\frac{1}{\sigma^{2}_{Z_{i} }}  }$.

\noindent Therefore, substitute Equation \ref{XplusepsionInverse} into ${|\boldsymbol{\Sigma_{Y}-\Sigma_{Y(Z+\epsilon) }{\Sigma}^{-1}_{Z+\epsilon}\Sigma_{(Z+\epsilon) Y}}|}$, we can obtain:
\begin{equation}
\resizebox{.99\hsize}{!}{$
\begin{split}
&{|\boldsymbol{\Sigma_{Y}-\Sigma_{Y(Z+\epsilon) }{\Sigma}^{-1}_{Z+\epsilon}\Sigma_{(Z+\epsilon) Y}}|} \\
=&\left | \begin{bmatrix}
\sigma^{2}_{Y_{1}}  & ...  & 0\\
\vdots  &  \ddots  & \vdots  \\
0 & ... & \sigma^{2}_{Y_{k}}
\end{bmatrix} - \begin{bmatrix}
\mathrm{cov} (Y_{1},Z_{1}+\epsilon )  & ...  & 0\\
\vdots  &  \ddots  & \vdots  \\
0 & ... & \mathrm{cov} (Y_{k},Z_{k}+\epsilon )
\end{bmatrix} \boldsymbol{{\Sigma}_{Z+\epsilon}^{-1}} \begin{bmatrix}
\mathrm{cov} (Y_{1},Z_{1}+\epsilon )  & ...  & 0\\
\vdots  &  \ddots  & \vdots  \\
0 & ... & \mathrm{cov} (Y_{k},Z_{k}+\epsilon )
\end{bmatrix}\right |  \\
=& \left | \begin{bmatrix}
\sigma^{2}_{Y_{1}} -\mathrm{cov}^{2}(Y_{1},Z_{1}+\epsilon) (\frac{ 1 }{\sigma^{2}_{Z_{1}}}-\frac{\lambda}{\sigma^{4}_{Z_{1}}})   & ...  &\mathrm{cov}(Y_{1},Z_{1}+\epsilon)\mathrm{cov}(Y_{k},Z_{k}+\epsilon)\frac{\lambda }{\sigma^{2}_{Z_{1}}\sigma^{2}_{Z_{k}}}\\
  \vdots &   &  \vdots\\
\mathrm{cov}(Y_{k},Z_{k}+\epsilon)\mathrm{cov}(Y_{1},Z_{1}+\epsilon)\frac{\lambda}{\sigma^{2}_{Z_{k}}\sigma^{2}_{Z_{1}}}   & ...  & \sigma^{2}_{Y_{k}} -\mathrm{cov}^{2}(Y_{k},Z_{k}+\epsilon)(\frac{ 1 }{\sigma^{2}_{Z_{k}}} -\frac{\lambda}{\sigma^{4}_{Z_{k}}}) \\
\end{bmatrix} \right |  \\
=& \left |\begin{bmatrix}
\sigma^{2}_{Y_{1}} -\frac{ 1 }{\sigma^{2}_{Z_{1}}} \mathrm{cov}^{2}(Y_{1},Z_{1})  & &\\
   &  \ddots  &  \\
   &   & \sigma^{2}_{Y_{k}} -\frac{ 1 }{\sigma^{2}_{Z_{k}}} \mathrm{cov}^{2}(Y_{k},Z_{k}) \\
\end{bmatrix} + \lambda \begin{bmatrix}
\frac{1}{\sigma^{4}_{Z_{1}}}\mathrm{cov}^{2}(Y_{1},Z_{1})    & ...  &\frac{1 }{\sigma^{2}_{Z_{1}}\sigma^{2}_{Z_{k}}}\mathrm{cov}(Y_{1},Z_{1})\mathrm{cov}(Y_{k},Z_{k})\\
  \vdots &   &  \vdots\\
\frac{1}{\sigma^{2}_{Z_{k}}\sigma^{2}_{Z_{1}}} \mathrm{cov}(Y_{k},Z_{k})\mathrm{cov}(Y_{1},Z_{1})  & ...  & \frac{1}{\sigma^{4}_{Z_{k}}}\mathrm{cov}^{2}(Y_{k},Z_{k})  \\
\end{bmatrix} \right | 
\end{split}
$}
\end{equation}
We use the notation $\boldsymbol{v}=
\begin{bmatrix}
  \frac{ 1 }{\sigma^{2}_{Z_{1}}} \mathrm{cov}(Y_{1},Z_{1}) & \cdots  &\frac{ 1 }{\sigma^{2}_{Z_{k}}} \mathrm{cov}(Y_{k},Z_{k})
\end{bmatrix} ^{T}$, and $\boldsymbol{V} = \mathrm{diag}(\frac{1}{\sigma^{2}_{Z_{1}}} \mathrm{cov}^{2}(Y_{1},Z_{1}), \cdots, \frac{1}{\sigma^{2}_{Z_{k}}} \mathrm{cov}^{2}(Y_{k},Z_{k}) )$. And utilize the rule of determinants of sums \cite{Marcus1990}, then we have:
\begin{equation} \label{determainatSum}
\begin{split}
|\boldsymbol{\Sigma_{Y}-\Sigma_{Y(Z+\epsilon) }{\Sigma}^{-1}_{Z+\epsilon}\Sigma_{(Z+\epsilon) Y}}|=&|(\boldsymbol{\Sigma_{Y}-V})+\lambda \boldsymbol{v}\boldsymbol{v}^{T}| \\
=& |\boldsymbol{\Sigma_{Y}-V}|+\lambda\boldsymbol{v}^{T}\boldsymbol{(\boldsymbol{\Sigma_{Y}-V})^{*}}\boldsymbol{v}
\end{split}
\end{equation} 
where $(\boldsymbol{\Sigma_{Y}-V})^{*}$ is the adjoint of the matrix $(\boldsymbol{\Sigma_{Y}-V})$.
For simplicity, we can rewrite ${|\boldsymbol{\Sigma_{Y}-\Sigma_{Y(Z+\epsilon) }{\Sigma}^{-1}_{Z+\epsilon}\Sigma_{(Z+\epsilon) Y}}|} $ as:
\begin{equation} \label{GauEquNominator}
\begin{split}
   & {|\boldsymbol{\Sigma_{Y}-\Sigma_{Y(Z+\epsilon) }{\Sigma}^{-1}_{Z+\epsilon}\Sigma_{(Z+\epsilon) Y}}|} \\
= &  \prod_{i=1}^{k}( \sigma^{2}_{Y_{i}} -\mathrm{cov}^{2}(Y_{i},Z_{i}) \frac{ 1 }{\sigma^{2}_{Z_{i}}}) + \Omega 
\end{split}
\end{equation}
where $\Omega = \lambda\boldsymbol{v}^{T}\boldsymbol{(\boldsymbol{\Sigma_{Y}-V})}^{*}\boldsymbol{v}$. The specific value of $\Omega$ can be obtained as:
\begin{equation} \label{Omega}
\begin{split}
 \Omega = \lambda \begin{bmatrix}
  \frac{ 1 }{\sigma^{2}_{Z_{1}}} \mathrm{cov}(Y_{1},Z_{1}) & \cdots  &\frac{ 1 }{\sigma^{2}_{Z_{k}}} \mathrm{cov}(Y_{k},Z_{k})
\end{bmatrix} \begin{bmatrix}
 V_{11} &  & \\
  & \ddots  & \\
  &  &V_{kk}
\end{bmatrix} \begin{bmatrix}
\frac{ 1 }{\sigma^{2}_{Z_{1}}} \mathrm{cov}(Y_{1},Z_{1}) \\
\vdots  \\
\frac{ 1 }{\sigma^{2}_{Z_{k}}} \mathrm{cov}(Y_{k},Z_{k})
\end{bmatrix}
\end{split}
\end{equation}
where the elements $V_{ii}, i \in [1,k]$  are minors of the matrix and expressed as:
\begin{equation} 
    V_{ii} = \prod_{j=1,j\ne i}^{k} \left [ \sigma^{2}_{Y_{j}}-\frac{1}{\sigma^{2}_{Z_{j}}}\mathrm{cov}^{2}(Z_{j},Y_{j})   \right ] 
\end{equation}
After some necessary steps, Equation \ref{Omega} is reduced to:
\begin{equation} \label{specificOmega}
\begin{split}
    \Omega = &\lambda \sum_{i=1}^{k} \frac{ \frac{ 1 }{\sigma^{4}_{Z_{i}}} \mathrm{cov}^{2}(Y_{i},Z_{i}) \prod_{j=1}^{k}( \sigma^{2}_{Y_{j}} -\mathrm{cov}^{2}(Y_{j},Z_{j}) \frac{ 1 }{\sigma^{2}_{Z_{j}}}) }{( \sigma^{2}_{Y_{i}} -\mathrm{cov}^{2}(Y_{i},Z_{i}) \frac{ 1 }{\sigma^{2}_{Z_{i}}})}  \\
     = & \lambda \prod_{i=1}^{k}( \sigma^{2}_{Y_{i}} -\mathrm{cov}^{2}(Y_{i},Z_{i}) \frac{ 1 }{\sigma^{2}_{Z_{i}}}) \cdot  \sum_{i=1}^{k} \frac{\mathrm{cov}^{2}(Z_{i},Y_{i})} {\sigma^{2}_{Z_{i}}(\sigma^{2}_{Z_{i}}\sigma^{2}_{Y_{i}}-\mathrm{cov}^{2}(Z_{i},Y_{i}))} \\
\end{split}
\end{equation}
Substitute Equation \ref{specificOmega} into Equation \ref{GauEquNominator}, we can get:
\begin{equation}
\begin{split}
   & {|\boldsymbol{\Sigma_{Y}-\Sigma_{Y(Z+\epsilon) }{\Sigma}^{-1}_{Z+\epsilon}\Sigma_{(Z+\epsilon) Y}}|}  \\
   =& \prod_{i=1}^{k}( \sigma^{2}_{Y_{i}} -\mathrm{cov}^{2}(Y_{i},Z_{i}) \frac{ 1 }{\sigma^{2}_{Z_{i}}})\cdot  (1+\lambda \sum_{i=1}^{k} \frac{\mathrm{cov}^{2}(Z_{i},Y_{i})} { \sigma^{2}_{Z_{i}}(\sigma^{2}_{Z_{i}}\sigma^{2}_{Y_{i}}-\mathrm{cov}^{2}(Z_{i},Y_{i}))} )
\end{split}
\end{equation}
Accordingly, $|\boldsymbol{\Sigma_{Y}-\Sigma_{YZ}{\Sigma}^{-1}_{Z}\Sigma_{ZY}}|$ is:
\begin{equation}
|\boldsymbol{\Sigma_{Y}-\Sigma_{YZ}{\Sigma}^{-1}_{Z}\Sigma_{ZY}}|=\prod_{i=1}^{k} (\sigma^{2}_{Y_{i}}-\frac{1}{\sigma^{2}_{Z_{i}}} \mathrm{cov}^{2}(Z_{i},Y_{i}) )
\end{equation}
As a result, $ \frac{|\boldsymbol{\Sigma_{Y|Z+\epsilon}}|}{|\boldsymbol{\Sigma_{Y|Z}}|}$ is expressed as:
\begin{equation} \label{GNConclusion}
\frac{|\boldsymbol{\Sigma_{Y|Z}}|}{|\boldsymbol{\Sigma_{Y|Z+\epsilon}}|} = \frac{\prod_{i=1}^{k}(\sigma^{2}_{Y_{i}}-\frac{1}{\sigma^{2}_{Z_{i}}} \mathrm{cov}^{2}(Z_{i},Y_{i}))}{\prod_{i=1}^{k}( \sigma^{2}_{Y_{i}} -\mathrm{cov}^{2}(Y_{i},Z_{i}) \frac{ 1 }{\sigma^{2}_{Z_{i}}}) \cdot  (1+\lambda \sum_{i=1}^{k} \frac{\mathrm{cov}^{2}(Z_{i},Y_{i})} { \sigma^{2}_{Z_{i}}(\sigma^{2}_{Z_{i}}\sigma^{2}_{Y_{i}}-\mathrm{cov}^{2}(Z_{i},Y_{i}))} ) }  
\end{equation}
Combine Equations \ref{GNConclusion} and \ref{epsilonwithXcovariance} together, the entropy change is expressed as:
\begin{equation} \label{LatentEntropyChangeFinal}
\begin{split}
\bigtriangleup S(\mathcal{T},\boldsymbol{\epsilon } )
 =&\frac{1}{2} \log  \frac{ 1}{(1+\sigma^{2}_{\epsilon } {\textstyle \sum_{i=1}^{k}\frac{1}{\sigma^{2}_{Z_{i} }} } )(1+\lambda {\sum_{i=1}^{k} \frac{\mathrm{cov}^{2}(Z_{i},Y_{i})} { \sigma^{2}_{Z_{i}}(\sigma^{2}_{Z_{i}}\sigma^{2}_{Y_{i}}-\mathrm{cov}^{2}(Z_{i},Y_{i}))}})}
\end{split}
\end{equation}
It is difficult to tell whether Equation \ref{LatentEntropyChangeFinal} is greater or smaller than $0$ directly. But one thing is for sure that when there is no Gaussian noise, Equation \ref{LatentEntropyChangeFinal} equals 0. However, we can use another way to compare the numerator and denominator in Equation \ref{LatentEntropyChangeFinal}.
Instead, we use the symbol $M$ to compare the numerator and denominator using subtraction. Let:
\begin{equation} 
\begin{split}
M=&1-(1+\sigma^{2}_{\epsilon } {\textstyle \sum_{i=1}^{k}\frac{1}{\sigma^{2}_{Z_{i} }} } )(1+\lambda {\sum_{i=1}^{k} \frac{\mathrm{cov}^{2}(Z_{i},Y_{i})} { \sigma^{2}_{Z_{i}}(\sigma^{2}_{Z_{i}}\sigma^{2}_{Y_{i}}-\mathrm{cov}^{2}(Z_{i},Y_{i}))}}) 
\end{split}
\end{equation}
Obviously, the variance $\sigma_{\epsilon}^{2}$ of the Gaussian noise control the result of $M$, while the mean $\mu_{\epsilon}$ has no influence. When $\sigma_{\epsilon}$ approaching 0, we have:
\begin{equation} 
\lim_{\sigma_{\epsilon }^{2} \to 0} M=0
\end{equation}
To determine if Gaussian noise can be positive noise, we need to determine whether the entropy change is large or smaller than 0.
\begin{equation} \label{LatentGN_Conlusion}
\left\{\begin{matrix}
 \bigtriangleup S(\mathcal{T},\boldsymbol{\epsilon } ) >0 & \mathrm{if} \  M > 0\\
\bigtriangleup S(\mathcal{T},\boldsymbol{\epsilon } ) \leq 0  & \mathrm{if}  \ M \leq 0
\end{matrix}\right.
\end{equation}
From the above equations, the sign of the entropy change is determined by the statistical properties of the embeddings and labels. Since ${\epsilon }^{2}\ge 0$ and $\lambda \ge 0$, we need to have a deep dive into the residual part, i.e., 
\begin{equation} \label{GNresidualpart}
{\sum_{i=1}^{k} \frac{\mathrm{cov}^{2}(Z_{i},Y_{i})} { \sigma^{2}_{Z_{i}}(\sigma^{2}_{Z_{i}}\sigma^{2}_{Y_{i}}-\mathrm{cov}^{2}(Z_{i},Y_{i}))}}=\sum_{i=1}^{k} \frac{\mathrm{cov}^{2}(Z_{i},Y_{i})} {\sigma^{4}_{Z_{i}}\sigma^{2}_{Y_{i}}(1-\rho_{Z_{i}Y_{i}}^2 )}
\end{equation}
where $\rho_{Z_{i}Y_{i}}$ is the correlation coefficient, and $\rho^{2}_{Z_{i}Y_{i}} \in [0,1]$. Eq. \ref{GNresidualpart} is greater than 0, As a result, the sign of the entropy change in the Gaussian noise case is less than zero. We can conclude that Gaussian noise added to the latent space is harmful to the task.

\subsection{Add Gaussian Noise to Raw Images}
Assuming that the pixels of the raw images follow a Gaussian distribution. The variation of task complexity by adding Gaussian noise to raw images can be formulated as:
\begin{equation} \label{GaussianNoiseInput}
\begin{split}
\bigtriangleup S(\mathcal{T},\boldsymbol{\epsilon } )=&H_{\mathcal{T}}(\boldsymbol{X}) - H_{\mathcal{T}}(\boldsymbol{X+\epsilon}) \\
= &\frac{1}{2} \log |\boldsymbol{\Sigma_{Y|X}}| -\frac{1}{2} \log|\boldsymbol{\Sigma_{Y|X+\epsilon}}| \\
=&\frac{1}{2} \log \frac{|\boldsymbol{\Sigma_{Y|X}}|}{|\boldsymbol{\Sigma_{Y|X+\epsilon}}|} \\
=&\frac{1}{2} \log \frac{|\boldsymbol{\Sigma_{Y}-\Sigma_{YX}{\Sigma}^{-1}_{X}\Sigma_{XY}}|}{|\boldsymbol{\Sigma_{Y}-\Sigma_{Y(X+\epsilon) }{\Sigma}^{-1}_{X+\epsilon}\Sigma_{(X+\epsilon) Y}}|} \\
=& \frac{1}{2} \log \frac{|\boldsymbol{\Sigma_{Y}-\Sigma_{YX}{\Sigma}^{-1}_{X}\Sigma_{XY}}|}{|\boldsymbol{\Sigma_{Y}-\Sigma_{YX }{\Sigma}^{-1}_{X+\epsilon}\Sigma_{X Y}}|}
\end{split}
\end{equation}
Borrowing the equations from the case of Gaussian noise added to the latent space, we have:
\begin{equation} \label{GaussianNoiseInputFinal}
\bigtriangleup S(\mathcal{T},\boldsymbol{\epsilon } )=\frac{1}{2} \log  \frac{ 1}{1+\lambda \sum_{i=1}^{k} \frac{\mathrm{cov}^{2}(X_{i},Y_{i})} { \sigma^{2}_{X_{i}}(\sigma^{2}_{X_{i}}\sigma^{2}_{Y_{i}}-\mathrm{cov}^{2}(X_{i},Y_{i}))}}
\end{equation}
Clearly, the introduction of Gaussian noise to each pixel in the original images has a detrimental impact on the task. \textbf{Note} that some studies have empirically shown that adding Gaussian noise to partial pixels of input images may be beneficial to the learning task \cite{Li2022Positive} \cite{Zhang2023Positive}.
%%%%%%%%%%%%%%%%%%%%%%%%%%%%%%%%%%%%%%%%%%%%%%%%%%%%%%%%%%%%

\section{Impact of Linear Transform Noise on Task Entropy}
\label{LinearTransformCase}
In our work, concerning the image level perspective, "linear transform noise" denotes an image that is perturbed by another image or a combination of other images. From the viewpoint of embeddings, "linear transform noise" refers to an embedding perturbed by another embedding or the combination of other embeddings.

\subsection{Inject Linear Transform Noise in Latent Space}
The entropy change of injecting linear transform noise into embeddings can be formulated as:
\begin{equation} \label{LinearNosieFormulaAppendix}
\begin{split}
\bigtriangleup S(\mathcal{T},Q\boldsymbol{Z})
= & H_{\mathcal{T}}(\boldsymbol{Z}) - H_{\mathcal{T}}(\boldsymbol{Z}+Q\boldsymbol{Z}) \\
= &H(\boldsymbol{Y},\boldsymbol{Z}) - H(\boldsymbol{Z}) -( H(\boldsymbol{Y},\boldsymbol{Z}+Q\boldsymbol{Z})- H(\boldsymbol{Z} ))\\
=&H(\boldsymbol{Y},\boldsymbol{Z}) - H(\boldsymbol{Y},\boldsymbol{Z}+Q\boldsymbol{Z})\\
=&\frac{1}{2} \log \frac{|\boldsymbol{\Sigma}_{\boldsymbol{Z}}||\boldsymbol{\Sigma_{Y}}-\boldsymbol{\Sigma_{YZ}}{\boldsymbol{\Sigma_{Z}^{-1}}}\boldsymbol{\Sigma_{ZY}}|}{|\boldsymbol{\Sigma}_{(I+Q)\boldsymbol{Z}}||\boldsymbol{\Sigma_{Y}-\Sigma_{YZ}}\boldsymbol{\Sigma^{-1}_{Z}}\boldsymbol{\Sigma_{ZY}}|} \\
=&\frac{1}{2} \log \frac{ 1 }{|I+Q|^{2}}\\
=&-\log|I+Q|
\end{split}
\end{equation}
Since we want the entropy change to be greater than 0, we can formulate Equation \ref{LinearNosieFormulaAppendix} as an optimization problem:
\begin{equation} \label{LinearNosieOptimizationAppendix}
\begin{split}
&   \max_{Q} \bigtriangleup S(\mathcal{T},Q\boldsymbol{Z}) \\
&s.t. \ rank(I+Q)=k \\
& \ \ \quad Q \sim I \\
&\ \quad  \left [ I+Q \right ]_{ii} \ge  \left [ I+Q \right ]_{ij}, i \ne j \\
& \ \quad  \left \| \left [ I+Q \right ]_{i}   \right \|_{1} = 1 
\end{split}
\end{equation}
where $\sim$ means the row equivalence.
The key to determining whether the linear transform is positive noise or not lies in the matrix of $Q$. The most important step is to ensure that $I+Q$ is invertible, which is $|(I+Q)| \ne 0$. For this, we need to investigate what leads $I+Q$ to be rank-deficient. The third constraint is to make the trained classifier get enough information about a specific embedding of an image and correctly predict the corresponding label. For instance, when an embedding $Z_{1}$ is perturbed by another embedding $Z_{2}$, the classifier predominantly relies on the information from $Z_{1}$ to predict the label $Y_{1}$. Conversely, if the perturbed embedding $Z_{2}$ takes precedence, the classifier struggles to accurately predict the label $Y_{1}$ and is more likely to predict it as label $Y_{2}$. The fourth constraint is the normalization of latent representations. 

\noindent \textbf{Rank Deficiency Cases} To avoid causing a rank deficiency of $I+Q$, we need to figure out the conditions that lead to rank deficiency. Here we show a simple case causing the rank deficiency. When the matrix $Q$ is a backward identity matrix \cite{Horn2012}, 
\begin{equation}
Q_{i,j}=\left\{\begin{matrix}
1,  & i+j=k+1\\
0,  & i+j\ne k+1
\end{matrix}\right.
\end{equation}
i.e.,
\begin{equation}
Q=\begin{bmatrix}
 0 & 0  & ... & 0 & 0 &1 \\
 0 & 0  & ... & 0 & 1 &0\\
 \vdots  &  \vdots&  & \vdots& \vdots & \vdots \\
0 & 1  & ... & 0 & 0 &0\\
1 & 0  & ... & 0 & 0 &0\\
\end{bmatrix}
\end{equation}
then $(I+Q) $ will be:
\begin{equation}
I+Q=\begin{bmatrix}
 1 & 0  & ... & 0 & 0 &1 \\
 0 & 1  & ... & 0 & 1 &0\\
 \vdots  &  \vdots&  & \vdots& \vdots & \vdots \\
0 & 1  & ... & 0 & 1 &0\\
1 & 0  & ... & 0 & 0 &1\\
\end{bmatrix}
\end{equation}
Thus, $I+Q$ will be rank-deficient when $Q$ is a backward identity. In fact, when the following constraints are satisfied, the $I+Q$ will be rank-deficient:
\begin{align} \label{Illcases}
\mathrm{HermiteForm} (I+Q)_{i}= \boldsymbol{0}, \quad  \exists i \in  \left [ 1,k \right ] 
\end{align}
where index $i$ is the row index, in this paper, the row index starts from $1$, and $\mathrm{HermiteForm}$ is the Hermite normal form \cite{Kannan1979}. 

\noindent \textbf{Full Rank Cases} Except for the rank deficiency cases, $I+Q$ has full rank and is invertible. Since $Q$ is a row equivalent to the identity matrix, we need to introduce the three types of elementary row operations as follows \cite{Shores2007}.
\begin{itemize}
\item [$\triangleright$ 1] \textbf{Row Swap} Exchange rows. \\
Row swap here allows exchanging any number of rows. This is slightly different from the original one that only allows any two-row exchange since following the original row swap will lead to a rank deficiency. When the $Q$ is derived from $I$ with \textbf{Row Swap}, it will break the third constraint. Therefore, \textbf{Row Swap} merely is considered harmful and would degrade the performance of deep models.
\item [$\triangleright$ 2] \textbf{Scalar Multiplication} Multiply any row by a constant $\beta$. This breaks the fourth constraint, thus degrading the performance of deep models. 
\item [$\triangleright$ 3] \textbf{Row Sum} Add a multiple of one row to another row. Then the matrix $I+Q$ would be like:
\begin{equation}
\begin{split}
 I+Q =& \begin{bmatrix}
 1 &  &  &  &\\
  & . &  & &\\
  &  & . & &\\
  &  &  & .& \\
  &  &  & &1
\end{bmatrix} + \begin{bmatrix}
 1 &  &  &  &\\
  & . &  &\beta  &\\
  &  & . & &\\
  &  &  & .& \\
  &  &  & &1
\end{bmatrix}  \\
 =&
\begin{bmatrix} 
2 &  &  &  &\\
  & . & \beta & &\\
  &  & . & &\\
  &  &  & .& \\
  &  &  & &2
\end{bmatrix}
\end{split}
\end{equation}
where $\beta$ can be at a random position beside the diagonal.
As we can see from the simple example, \textbf{Row Sum} breaks the fourth constraint and makes entropy change smaller than 0.
\end{itemize}
From the above discussion, none of the single elementary row operations can guarantee positive effects on deep models. 

\noindent However, if we combine the elementary row operations, it is possible to make the entropy change greater than 0 as well as satisfy the constraints. For example, we combine the \textbf{Row Sum} and \textbf{Scalar Multiplication} to generate the $Q$: 
\begin{equation}
\begin{split}
I+Q =&\begin{bmatrix}
 1 &  &  &  &\\
  & . &  & &\\
  &  & . & &\\
  &  &  & .& \\
  &  &  & & 1
\end{bmatrix} + \begin{bmatrix}
 -0.5 & 0.5 &  &  &\\
  & . & .  & &\\
  &  & . &. &\\
  &  &  & .&0.5 \\
 0.5 &  &  & & -0.5
\end{bmatrix} \\
=&\begin{bmatrix}
 0.5 & 0.5 &  &  &\\
  & . & .  & &\\
  &  & . &. &\\
  &  &  & .&0.5 \\
 0.5 &  &  & & 0.5
\end{bmatrix}
\end{split}
\end{equation}
In this case, $\bigtriangleup S(\mathcal{T},Q\boldsymbol{X}) > 0$ when $Q = -0.5I$. The constraints are satisfied. This is just a simple case of adding linear transform noise that benefits deep models. Actually, there exists a design space of $Q$ that within the design space, deep models can reduce task entropy by injecting linear transform noise into the embeddings. To this end, we demonstrate that linear transform can be positive noise.

\noindent From the discussion in this section, we can draw conclusions that \textbf{Linear Transform Noise} can be positive under certain conditions, while \textbf{Gaussian Noise} and \textbf{Salt-and-pepper Noise} are harmful noise. From the above analysis, the conditions that satisfy positive noise form a design space. Exploring the design space of positive noise is an important topic for future work.

\subsubsection{Optimal Quality Matrix of Linear Transform Noise}
\label{OptimalQDerivation}
The optimal quality matrix should maximize the entropy change and therefore theoretically define the minimized task complexity. The optimization problem as formulated in Equation \ref{LinearNosieOptimization} is:
\begin{equation} 
\begin{split}
&   \max_{Q} -\log|I+Q|\\
&s.t. \ rank(I+Q)=k \\
& \ \ \quad Q \sim I \\
&\ \quad  \left [ I+Q \right ]_{ii} \ge  \left [ I+Q \right ]_{ij}, i \ne j \\
& \ \quad  \left \| \left [ I+Q \right ]_{i}   \right \|_{1} = 1 
\end{split}
\end{equation}
Maximizing the entropy change is to minimize the determinant of the matrix sum of $I$ and $Q$. A simple but straight way is to design the matrix $Q$ that makes the elements in $I+Q$ equal, i.e., 
\begin{equation}
    I+Q = \begin{bmatrix}
 1/k & \cdots  &  1/k\\
 \vdots  & \cdots & \vdots  \\
 1/k & \cdots  &  1/k
\end{bmatrix} 
\end{equation}
The determinant of the above equation is 0, but it breaks the first constraint of $\ rank(I+Q)=k$. However, by adding a small constant into the diagonal, and minus another constant by other elements, we can get:
\begin{equation}
    I+Q = \begin{bmatrix}
 1/k+c_{1}  & \cdots  & &  1/k-c_{2}\\
 1/k-c_{2}  & \ddots  & & \vdots  \\
\vdots  &  & \ddots & 1/k-c_{2}  \\
 1/k-c_{2} & \cdots  &  1/k-c_{2}  &1/k+c_{1}
\end{bmatrix} 
\end{equation}
Under the constraints, we can obtain the two constants that fulfill the requirements:
\begin{equation}
    c_{1}= \frac{k-1}{k(k+1)}, \quad c_{2} = \frac{1}{k(k+1)} 
\end{equation}
Therefore, the corresponding $Q$ is:
\begin{equation}\label{OptimalQAppendix}
Q_{optimal}=\mathrm{diag} \left ( \frac{1}{k+1}-1, \dots, \frac{1}{k+1}-1  \right ) +\frac{1}{k+1} \boldsymbol{1}_{k\times k}
\end{equation}
and the corresponding $I+Q$ is:
\begin{equation}
    I+Q = \begin{bmatrix}
 2/(k+1)  & \cdots  & &  1/(k+1)\\
 1/(k+1)  & \ddots  & & \vdots  \\
\vdots  &  & \ddots & 1/(k+1) \\
 1/(k+1) & \cdots  &  1/(k+1)  &2/(k+1)
\end{bmatrix} 
\end{equation}
As a result, the determinant of optimal $I+Q$ can be obtained by following the identical procedure as Equation \ref{determainatSum}:
\begin{equation}
\left | I+Q \right | = \frac{1}{(k+1)^{k-1} }  
\end{equation}
The upper boundary of entropy change of linear transform noise is determined: 
\begin{equation}
 \bigtriangleup S(\mathcal{T},Q\boldsymbol{X } )_{upper} = (k-1)\log{(k+1) } 
\end{equation}
%%%%%%%%%%%%%%%%%%%%%%%%%%%%%

\subsection{Add Linear Transform Noise to Raw Images}
\noindent In this case, the task entropy with linear transform noise can be formulated as:
\begin{equation}
\begin{split}
H_{\mathcal{T}}(\boldsymbol{X}+Q\boldsymbol{X})=&-\sum_{\boldsymbol{Y} \in \mathcal{Y}}p(\boldsymbol{Y}|\boldsymbol{X}+Q\boldsymbol{X})\log{p(\boldsymbol{Y}|\boldsymbol{X}+Q\boldsymbol{X})} \\
=&-\sum_{\boldsymbol{Y} \in \mathcal{Y}}p(\boldsymbol{Y}|(I+Q)\boldsymbol{X})\log{p(\boldsymbol{Y}|(I+Q)\boldsymbol{X})} 
\end{split}
\end{equation}
where $I$ is an identity matrix, and $Q$ is derived from $I$ using elementary row operations. Assuming that the pixels of the raw images follow a Gaussian distribution. The conditional distribution of $\boldsymbol{Y}$ given $\boldsymbol{X}+Q\boldsymbol{X}$ is also multivariate subjected to the normal distribution, which can be formulated as:
\begin{equation}
\boldsymbol{Y}|(I+Q)\boldsymbol{X}\sim \mathcal{N}(\mathbb{E}({\boldsymbol{Y}|(I+Q)\boldsymbol{X}}), var(\boldsymbol{Y}|(I+Q)\boldsymbol{X}))
\end{equation}
Since the linear transform matrix is invertible, applying the linear transform to $\boldsymbol{X}$ does not alter the distribution of the $\boldsymbol{X}$. It is straightforward to obtain:
\begin{equation}
\boldsymbol{\mu}_{\boldsymbol{Y}|(I+Q)\boldsymbol{X}}=\boldsymbol{\mu_{Y}}+\boldsymbol{\Sigma_{YX}}\boldsymbol{{\Sigma}^{-1}_{X}}(I+Q)^{-1}((I+Q)X-(I+Q)\boldsymbol{\mu_{X}})
\end{equation}
\begin{equation}
\boldsymbol{\Sigma}_{{(\boldsymbol{Y}|(I+Q)\boldsymbol{X}})}=\boldsymbol{\Sigma_{Y}-\Sigma_{YX}}\boldsymbol{\Sigma^{-1}_{X}}\boldsymbol{\Sigma_{XY}}
\end{equation}

\noindent Thus, the variation of task entropy adding linear transform noise can be formulated as:
\begin{equation} 
\begin{split}
\bigtriangleup S(\mathcal{T},Q\boldsymbol{X})=
&H_{\mathcal{T}}(\boldsymbol{X}) - H_{\mathcal{T}}(\boldsymbol{X}+Q\boldsymbol{X}) \\
= &\frac{1}{2} \log |\boldsymbol{\Sigma_{Y|X}}| -\frac{1}{2} \log|{\Sigma_{\boldsymbol{Y}|\boldsymbol{X}+Q\boldsymbol{X}}}| \\
=&\frac{1}{2} \log \frac{|\boldsymbol{\Sigma_{Y|X}}|}{|{\Sigma_{\boldsymbol{Y}|\boldsymbol{X}+Q\boldsymbol{X}}}|} \\
=&\frac{1}{2} \log \frac{|\boldsymbol{\Sigma_{Y}-\Sigma_{YX}{\Sigma}^{-1}_{X}\Sigma_{XY}}|}{|\boldsymbol{\Sigma_{Y}-\Sigma_{YX}}\boldsymbol{\Sigma^{-1}_{X}}\boldsymbol{\Sigma_{XY}}|} \\
=&0
\end{split}
\end{equation}
The entropy change of 0 indicates that the implementation of linear transformation to the raw images could not help reduce the complexity of the task.
 
\section{Influence of Salt-and-pepper Noise on Task Entropy}
\label{ImpulseCase}
\noindent Salt-and-pepper noise is a common type of noise that can occur in images due to various factors, such as signal transmission errors, faulty sensors, or other environmental factors \cite{Chan2005}. Salt-and-pepper noise is often considered to be an independent process because it is a type of random noise that affects individual pixels in an image independently of each other \cite{Gonzales2007}. 
\subsection{Inject Salt-and-pepper Noise in Latent Space}
\label{SaltandpepperNoiseLatentSpace}
The entropy change of injecting salt-and-pepper noise can be formulated as:
\begin{align*}
\bigtriangleup S(\mathcal{T},\boldsymbol{\epsilon } )
= & H_{\mathcal{T}}(\boldsymbol{Z}) - H_{\mathcal{T}}(\boldsymbol{Z}\boldsymbol{\epsilon})  \\
= & H(\boldsymbol{Y},\boldsymbol{Z}) - H(\boldsymbol{Y},\boldsymbol{Z\epsilon}) \\
= &H(\boldsymbol{Y},\boldsymbol{Z}) - H(\boldsymbol{Z}) -( H(\boldsymbol{Y},\boldsymbol{Z}\boldsymbol{\epsilon })- H(\boldsymbol{Z} ))\\
= & \mathbb{E}\!\left[\log\frac{1}{p(\boldsymbol{Z},\boldsymbol{Y})}\right] - \mathbb{E}\!\left[\log\frac{1}{p(\boldsymbol{Z\epsilon},\boldsymbol{Y})}\right] \\
\le & \mathbb{E}\!\left[\log\frac{1}{p(\boldsymbol{Z},\boldsymbol{Y})}\right] - \mathbb{E}\!\left[\log\frac{1}{p(\boldsymbol{Z},\boldsymbol{Y})}\right] \;=\; 0,
\end{align*}

The entropy change is smaller than 0, therefore, the salt-and-pepper is a pure detrimental noise to the learning task.

\subsection{Add Salt-and-pepper Noise to Raw Images}
The task entropy with salt-and-pepper noise is rewritten as:
\begin{equation}\label{impluse}
\begin{split}
H_{\mathcal{T}}(\boldsymbol{X}\boldsymbol{\epsilon})=&-\sum_{\boldsymbol{Y} \in \mathcal{Y}}p(\boldsymbol{Y}|\boldsymbol{X}\boldsymbol{\epsilon})\log{p(\boldsymbol{Y}|\boldsymbol{X}\boldsymbol{\epsilon})}
\end{split}
\end{equation}

\noindent Since $\boldsymbol{\epsilon}$ is independent of $\boldsymbol{X}$ and $\boldsymbol{Y}$, and $I(\boldsymbol{Y};\boldsymbol{X \epsilon})\le I(\boldsymbol{Y};\boldsymbol{X})\ \Longrightarrow \ H(\boldsymbol{Y}\mid \boldsymbol{X})\le H(\boldsymbol{Y}\mid \boldsymbol{X\epsilon})$. The entropy change satisfies:
\begin{equation}
\Delta S(\mathcal{T},\boldsymbol{\epsilon})
=H(\boldsymbol{Y} \mid \boldsymbol{X})-H(\boldsymbol{Y}\mid \boldsymbol{X\epsilon})\le 0
\end{equation}
Therefore, salt-and-pepper noise can not help reduce the complexity of the task, and therefore, it is considered a type of pure detrimental noise.

\noindent From the discussion in this section, we can draw conclusions that \textbf{Linear Transform Noise} can be positive under certain conditions, while \textbf{Gaussian Noise} and \textbf{Salt-and-pepper Noise} are harmful noise. From the above analysis, the conditions that satisfy positive noise are forming a design space. Exploring the positive noise space is an important topic for future work.

\begin{figure*}[!tp]
\begin{center}
\includegraphics[width=1.0\textwidth]{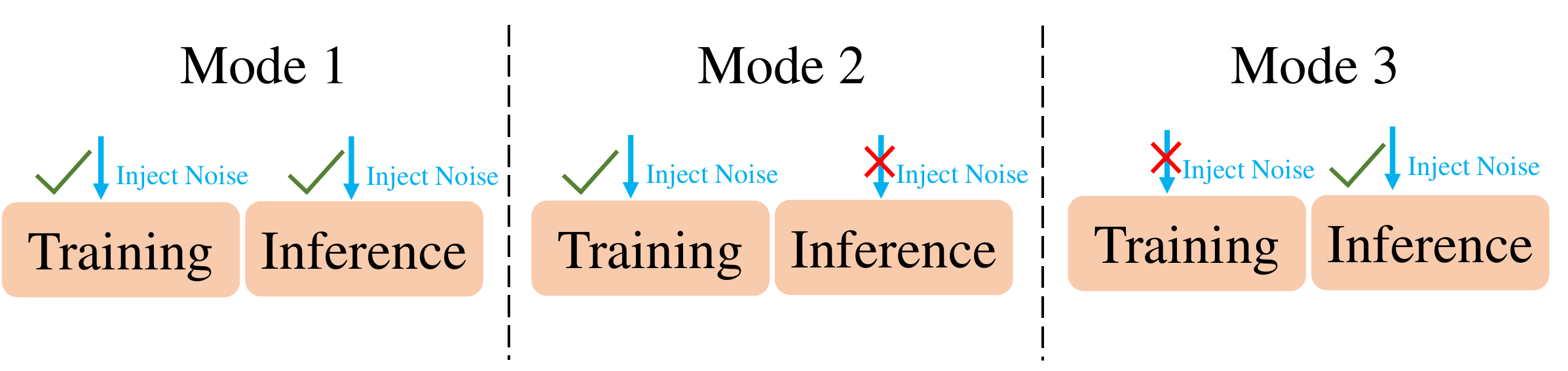}
\end{center}
\caption{Differnet Noise Injection Modes. The green tick mark means noise injection operations are implemented, while the red cross means no noise injection. We adopt the Mode 1 in this work. Modes 2 and 3 are for future exploration.}
\label{injectionmodes}
\end{figure*}

\section{Experimental Setting}
%The noise was added to both the training stage and the inference stage.
We introduce the implementation details in this part. There are three different noise injection modes, as shown in Fig.~\ref{injectionmodes}. We follow Mode 3 in this paper. Model details are shown in Table \ref{ResNetscale} and \ref{ViTscale}. The image resolution is $224 \times 224$ for all the experiments. Pre-trained models on ImageNet-21K are used as the backbone. We train all ResNet and ViT-based models using AdamW optimizer \cite{Loshchilov2017}. We set the learning rate of each parameter group using a cosine annealing schedule with a minimum of $1e-7$. The data augmentation for training only includes the random resized crop and normalization.

\begin{table}[]
\caption{ Details of ResNet Models. The columns "18-layer", "34-layer", "50-layer", and "101-layer" show the specifications of ResNet-18, ResNet-34, ResNet-50, and ResNet-101, separately. } 
\centering
\resizebox{\linewidth}{!}{
\begin{tabular}{cc|cccc}
\hline
\multicolumn{1}{c|}{Layer name} & Output size      & \multicolumn{1}{c|}{18-layer}               & \multicolumn{1}{c|}{34-layer}                  & \multicolumn{1}{c|}{50-layer}                                   & 101-layer                                                             \\ \hline
\multicolumn{1}{c|}{conv1}      & 112 $\times$ 112 & \multicolumn{4}{c}{7 $\times$ 7, 64, stride 2}                           \\ \hline
\multicolumn{1}{c|}{}           &                  & \multicolumn{4}{c}{3 $\times$ 3, max pool, stride 2}           \\ \cline{3-6} 
\multicolumn{1}{c|}{conv2\_x}   & 56 $\times$ 56   & \multicolumn{1}{c|}{\begin{tabular}[c]{@{}c@{}}$\begin{bmatrix} 3 \times 3  & 64\\   3 \times 3  & 64 \end{bmatrix} \times 2$\end{tabular}}   & \multicolumn{1}{c|}{\begin{tabular}[c]{@{}c@{}}$\begin{bmatrix} 3 \times 3  & 64\\  3 \times 3  & 64 \end{bmatrix} \times 3$\end{tabular}}   & \multicolumn{1}{c|}{\begin{tabular}[c]{@{}c@{}}$\begin{bmatrix} 1 \times 1  & 64\\  3 \times 3  & 64 \\  1 \times 1  & 256 \end{bmatrix} \times 3$\end{tabular}}    & \begin{tabular}[c]{@{}c@{}}$\begin{bmatrix} 1 \times 1  & 64\\   3 \times 3  & 64 \\  1 \times 1  & 256 \end{bmatrix} \times 3$\end{tabular}    \\ \hline
\multicolumn{1}{c|}{conv3\_x}   & 28 $\times$ 28   & \multicolumn{1}{c|}{\begin{tabular}[c]{@{}c@{}}$\begin{bmatrix} 3 \times 3  & 128\\   3 \times 3  & 128 \end{bmatrix} \times 2$\end{tabular}} & \multicolumn{1}{c|}{\begin{tabular}[c]{@{}c@{}}$\begin{bmatrix} 3 \times 3  & 128\\   3 \times 3  & 128 \end{bmatrix} \times 4$\end{tabular}} & \multicolumn{1}{c|}{\begin{tabular}[c]{@{}c@{}}$\begin{bmatrix} 1 \times 1  & 128\\\   3 \times 3  & 128 \\  1 \times 1  & 512 \end{bmatrix} \times 4$\end{tabular}}   & \begin{tabular}[c]{@{}c@{}}$\begin{bmatrix} 1 \times 1  & 128\\\   3 \times 3  & 128 \\  1 \times 1  & 512 \end{bmatrix} \times 4$\end{tabular}   \\ \hline
\multicolumn{1}{c|}{conv4\_x}   & 14 $\times$ 14   & \multicolumn{1}{c|}{\begin{tabular}[c]{@{}c@{}}$\begin{bmatrix} 3 \times 3  & 256\\   3 \times 3  & 256 \end{bmatrix} \times 2$\end{tabular}} & \multicolumn{1}{c|}{\begin{tabular}[c]{@{}c@{}}$\begin{bmatrix} 3 \times 3  & 256\\   3 \times 3  & 256 \end{bmatrix} \times 6$\end{tabular}} & \multicolumn{1}{c|}{\begin{tabular}[c]{@{}c@{}}$\begin{bmatrix} 1 \times 1  & 256\\   3 \times 3  & 256 \\  1 \times 1  & 1024 \end{bmatrix} \times 6$\end{tabular}} & \begin{tabular}[c]{@{}c@{}}$\begin{bmatrix} 1 \times 1  & 256\\\   3 \times 3  & 256 \\  1 \times 1  & 1024 \end{bmatrix} \times 23$\end{tabular} \\ \hline
\multicolumn{1}{c|}{conv5\_x}   & 7 $\times$ 7     & \multicolumn{1}{c|}{\begin{tabular}[c]{@{}c@{}}$\begin{bmatrix} 3 \times 3  & 512\\   3 \times 3  & 512 \end{bmatrix} \times 2$\end{tabular}} & \multicolumn{1}{c|}{\begin{tabular}[c]{@{}c@{}}$\begin{bmatrix} 3 \times 3  & 512\\   3 \times 3  & 512 \end{bmatrix} \times 3$\end{tabular}} & \multicolumn{1}{c|}{\begin{tabular}[c]{@{}c@{}}$\begin{bmatrix} 1 \times 1  & 512\\   3 \times 3  & 512 \\  1 \times 1  & 2048 \end{bmatrix} \times 3$\end{tabular}} & \begin{tabular}[c]{@{}c@{}}$\begin{bmatrix} 1 \times 1  & 512\\   3 \times 3  & 512 \\  1 \times 1  & 2048 \end{bmatrix} \times 3$\end{tabular} \\ \hline
\multicolumn{1}{c|}{}           & 1 $\times$ 1     & \multicolumn{4}{c}{average pool, 1000-d fc, softmax}       \\ \hline
\multicolumn{2}{c|}{Params}     & \multicolumn{1}{c|}{11M}                                  & \multicolumn{1}{c|}{22M}                             & \multicolumn{1}{c|}{26M}         &             45M     \\ \hline
\end{tabular}}
\label{ResNetscale}
\end{table}

\begin{table}[]
\caption{ Details of ViT Models. Each row shows the specifications of a kind of ViT model. ViT-T, ViT-S, ViT-B, and ViT-L represent ViT Tiny, ViT Small, ViT Base, and ViT Large, separately. } 
\centering
\begin{tabular}{cccccc}
\hline
ViT Model & Layers & Hidden size & MLP size & Heads & Params \\ \hline
ViT-T     & 12     & 192         & 768      & 3     & 5.7M   \\ 
ViT-S     & 12     & 384         & 1536     & 6     & 22M     \\ 
ViT-B     & 12     & 768         & 3072     & 12    & 86M    \\ 
ViT-L     & 12     & 1024        & 4096     & 16    & 307M   \\ \hline
\end{tabular}
\label{ViTscale}
\end{table}

\begin{table}[]
\caption{ Variants of ViT with different kinds of noise on TinyImageNet. Vanilla means the vanilla model without noise. Accuracy is shown in percentage. Gaussian noise used here is subjected to standard normal distribution. Linear transform noise used in this table is designed to be positive noise. The difference is shown in the bracket.} 
\centering
\begin{tabular}{ccccc}
\hline
Model  & DeiT & SwinTransformer & BeiT & ConViT \\ \hline
Vanilla      & 85.02 (+0.00)     & 90.84 (+0.00)    & 88.64  (+0.00)   & 90.69 (+0.00)     \\
+ Gaussian Noise         & 84.70 (-0.32)    & 90.34 (-0.50)     & 88.40 (-0.24)    & 90.40  (-0.29)    \\
+ Linear Transform Noise & \textbf{86.50 (+1.48)}     & \textbf{95.68 (+4.84)}    & \textbf{91.78 (+3.14)}    & \textbf{93.07 (+2.38)}   \\
+ Salt-and-pepper Noise  & 84.03 (-1.01)     & 87.12 (-3.72)    & 42.18 (-46.46)    & 89.93 (-0.76)   \\ \hline
 Params.  & 86M     & 87M    & 86M    & 86M  \\ \hline
\end{tabular}
\label{ViTVariantImageNet}
\end{table}

\begin{table}[] 
\caption{ ResNet with different kinds of noise on TinyImageNet. Vanilla means the vanilla model without noise. Accuracy is shown in percentage. Gaussian noise used here is subjected to standard normal distribution. Linear transform noise used in this table is designed to be positive noise. The difference is shown in the bracket.}
\centering
\begin{tabular}{ccccc}
\hline
Model   & ResNet-18 & ResNet-34 & ResNet-50 & ResNet-101 \\ \hline
Vanilla    & 64.01 (+0.00)     & 67.04 (+0.00)    & 69.47  (+0.00)   & 70.66 (+0.00)     \\
+ Gaussian Noise         & 63.23 (-0.78)    & 65.71 (-1.33)     & 68.17 (-1.30)    & 69.13  (-1.53)    \\
+ Linear Transform Noise & \textbf{73.32 (+9.31)}     & \textbf{76.70 (+9.66)}    & \textbf{76.88 (+7.41)}    & \textbf{77.30   (+6.64)}   \\
+ Salt-and-pepper Noise  & 55.97 (-8.04)     & 63.52 (-3.52)    & 49.42 (-20.25)    & 53.88   (-16.78)   \\ \hline
\end{tabular}
\label{ResNetTiny}
\end{table}

\begin{table}[h]
\caption{ ViT with different kinds of noise on TinyImageNet. Vanilla means the vanilla model without injecting noise. Accuracy is shown in percentage. Gaussian noise used here is subjected to standard normal distribution. Linear transform noise used in this table is designed to be positive noise. The difference is shown in the bracket. Note \textbf{ViT-L is overfitting on TinyImageNet} \cite{Dosovitskiy20} \cite{Steiner2021}.} 
\centering
\begin{tabular}{ccccc}
\hline
Model                       & ViT-T & ViT-S & ViT-B & ViT-L \\ \hline
Vanilla                     & 81.75 (+0.00)     & 86.78 (+0.00)    & 90.48  (+0.00)   & 93.32 (+0.00)     \\
+ Gaussian Noise         & 80.95 (-0.80)    & 85.66 (-1.12)     & 89.61 (-0.87)    & 92.31 (-1.01)    \\
+ Linear Transform Noise & \textbf{82.50 (+0.75)}     & \textbf{91.62 (+4.84)}    & \textbf{94.92 (+4.44)}    & \textbf{93.63 (+0.31)}   \\
+ Salt-and-pepper Noise  & 79.34 (-2.41)     & 84.66 (-2.12)    & 87.45 (-3.03)    & 83.48  (-9.84)   \\ \hline
\end{tabular}
\label{ViTTinyImageNet}
\end{table}

\noindent \textbf{CNN(ResNet) Setting} The training epoch is set to 100. We initialized the learning rate as $0$ and linearly increase it to $0.001$ after 10 warmup steps. All the experiments of CNNs are trained on a single Tesla V100 GPU with 32 GB. The batch size for ResNet18, ResNet34, ResNet50, and ResNet101 are 1024, 512, 256, and 128, respectively.

\noindent \textbf{ViT and Variants Setting}
All the experiments of ViT and its variants are trained on a single machine with 8 Tesla V100 GPUs. For vanilla ViTs, including ViT-T, ViT-S, ViT-B, and ViT-L, the training epoch is set to 50 and the input patch size is $16\times16$. We initialized the learning rate as 0 and linearly increase it to $0.0001$ after 10 warmup steps. We then decrease it by the cosine decay strategy. For experiments on the variants of ViT, the training epoch is set to 100 and the learning rate is set to $0.0005$ with 10 warmup steps.

\section{ More Experiment Results}
We show more experiment results of injecting positive noise to other variants of the ViT family, such as SwinTransformer, DeiT, ConViT, and BeiT, and implement them on the smaller dataset, i.e., TinyImageNet. Note, considering limited computational resources, all the experiments in the supplementary are on the TinyImageNet. The strength of positive noise is set to 0.3. The noise is injected into the last layer.

\subsection{Inject Positive Noise to Variants of ViT}
As demonstrated in the paper, the positive noise can be injected into the ViT family. Therefore, in this section, we explore the influence of positive noise on the variants of the ViT. The positive noise used here is identical to that in the paper. For this, we comprehensively compare noise injection to ConViT \cite{d2021convit}, BeiT \cite{beit}, DeiT \cite{Touvron2021DeiT}, and Swin Transformer \cite{liu2021Swin}, and comparisons results are reported in Tabel \ref{ViTVariantImageNet}. As expected, these variants of ViTs get benefit from the positive noise. The additional four ViT variants are at the base scale, whose parameters are listed in the table's last row. For a fair comparison, we use identical experimental settings for each kind of experiment. For example, we use the identical setting for vanilla ConViT, ConViT with different kinds of noise. From the experimental results, we can observe that the different variants of ViT benefit from positive noise and significantly improve prediction accuracy. The results on different scale datasets and variants of the ViT family demonstrate that positive noise can universally improve the model performance by a wide margin. 

\subsection{Positive Noise on TinyImageNet}
We also implement experiments of ResNet and ViT on the smaller dataset TinyImageNet, and the results are shown in Table \ref{ResNetTiny} and \ref{ViTTinyImageNet}. As shown in the tables, positive noise also benefits the deep models on the small dataset. From the experiment results of CNN and ViT family on ImageNet and TinyImageNet, we can find that the positive noise has better effects on larger datasets than smaller ones. This makes sense because as shown in the section on optimal quality matrix, the upper boundary of the entropy change is determined by the size, i.e., the number of data samples, of the dataset, smaller datasets have less number of data samples, which means the upper boundary of the small datasets is lower than the large datasets. Therefore, the positive noise of linear transform noise has better influences on large than small datasets.

\begin{table*}[t]
\caption{ Comparison with SOTA methods on \textbf{Office-Home}. The best performance is marked in red.} 
\centering
\setlength{\tabcolsep}{0.1mm}{} 
\begin{tabular}{ cccccccccccccc }
\hline
Method        & Ar2Cl & Ar2Pr & Ar2Re & Cl2Ar & Cl2Pr & Cl2Re & Pr2Ar & Pr2Cl & Pr2Re & Re2Ar & Re2Cl & Re2Pr & Avg. \\ \hline
ResNet-50\cite{Kaiming16}     & 44.9                & 66.3                & 74.3                & 51.8                & 61.9                & 63.6                & 52.4                & 39.1                & 71.2                & 63.8                & 45.9                & 77.2                & 59.4 \\ 
MinEnt\cite{Grandvalet04}        & 51.0                & 71.9                & 77.1                & 61.2                & 69.1                & 70.1                & 59.3                & 48.7                & 77.0                & 70.4                & 53.0                & 81.0                & 65.8 \\ 
SAFN\cite{Xu19}          & 52.0                & 71.7                & 76.3                & 64.2                & 69.9                & 71.9                & 63.7                & 51.4                & 77.1                & 70.9                & 57.1                & 81.5                & 67.3 \\ 
CDAN+E\cite{Long18}        & 54.6                & 74.1                & 78.1                & 63.0                & 72.2                & 74.1                & 61.6                & 52.3                & 79.1                & 72.3                & 57.3                & 82.8                & 68.5 \\ 
DCAN\cite{LiAAAI20}          & 54.5                & 75.7                & 81.2                & 67.4                & 74.0                & 76.3                & 67.4                & 52.7                & 80.6                & 74.1                & 59.1                & 83.5                & 70.5 \\ 
BNM \cite{CuiCVPR20}          & 56.7                & 77.5                & 81.0                & 67.3                & 76.3                & 77.1                & 65.3                & 55.1                & 82.0                & 73.6                & 57.0                & 84.3                & 71.1 \\ 
SHOT\cite{Liang20}          & 57.1                & 78.1                & 81.5                & 68.0                & 78.2                & 78.1                & 67.4                & 54.9                & 82.2                & 73.3                & 58.8                & 84.3                & 71.8 \\ 
ATDOC-NA\cite{LiangCVPR21}      & 58.3                & 78.8                & 82.3                & 69.4                & 78.2                & 78.2                & 67.1                & 56.0                & 82.7                & 72.0                & 58.2                & 85.5                & 72.2 \\ \hline
ViT-B\cite{Dosovitskiy20}         & 54.7                & 83.0                & 87.2                & 77.3                & 83.4                & 85.6                & 74.4                & 50.9                & 87.2                & 79.6                & 54.8                & 88.8                & 75.5 \\
TVT-B\cite{Yang23}         & 74.9                & 86.8                & 89.5                & 82.8                & 88.0                & 88.3                & 79.8                & 71.9                & 90.1                & 85.5                & 74.6                & 90.6                & 83.6 \\ 
CDTrans-B\cite{Xu22}     & 68.8                & 85.0                & 86.9                & 81.5                & 87.1                & 87.3                & 79.6                & 63.3                & 88.2                & 82.0                & 66.0                & 90.6                & 80.5 \\ 
SSRT-B \cite{Sun2022CVPR}       & 75.2                & 89.0                & 91.1                & 85.1                & 88.3                & 90.0                & 85.0                & 74.2                & 91.3                & 85.7                & 78.6                & 91.8                & 85.4 \\
NoisyTVT-B &  {\color{red}\textbf{78.3}}   &  {\color{red}\textbf{90.6}}     & {\color{red}\textbf{91.9}}      &  {\color{red}\textbf{87.8}}       &  {\color{red}\textbf{92.1}}                & {\color{red}\textbf{91.9}}     & {\color{red}\textbf{85.8}}    & {\color{red}\textbf{78.7}}             & {\color{red}\textbf{93.0}}   &  {\color{red}\textbf{88.6}}   &  {\color{red}\textbf{80.6}}                &  {\color{red}\textbf{93.5}}    & {\color{red}\textbf{87.7}} \\ \hline
\end{tabular}
\label{Office-HomeAppendix}
\end{table*}

\begin{table*}[t]
\caption{ Comparison with SOTA methods on \textbf{Visda2017}. The best performance is marked in red.}
\centering
\setlength{\tabcolsep}{0.7mm}{} 
\begin{tabular}{ cccccccccccccc }
\hline
Method     & plane & bcycl & bus & car & horse & knife & mcycl & person & plant & sktbrd & train & truck & Avg. \\ \hline
ResNet-50\cite{Kaiming16}     & 55.1                & 53.3              & 61.9              & 59.1               &80.6                &17.9             &  79.7               & 31.2                &  81.0               & 26.5                &  73.5               & 8.5                & 52.4 \\ 
DANN\cite{Ganin15}          & 81.9              & 77.7                &  82.8                & 44.3                &  81.2                & 29.5                &  65.1                & 28.6                &  51.9              & 54.6                & 82.8                &7.8                & 57.4\\ 
MinEnt\cite{Grandvalet04}        & 80.3               & 75.5                & 75.8               &48.3                &  77.9                & 27.3                & 69.7                & 40.2               & 46.5                & 46.6               & 79.3                &16.0                & 57.0 \\ 
SAFN\cite{Xu19}          &93.6                & 61.3                & 84.1               & 70.6               & 94.1               & 79.0               &  91.8                & 79.6               &  89.9                & 55.6                &  89.0              & 24.4              & 76.1 \\ 
CDAN+E\cite{Long18}        & 85.2               &  66.9              & 83.0                &  50.8              & 84.2                & 74.9                & 88.1               &74.5                & 83.4             &76.0              &81.9              &  38.0                & 73.9 \\ 
BNM \cite{CuiCVPR20}          &89.6            & 61.5                &  76.9                & 55.0             & 89.3             & 69.1                & 81.3             & 65.5            &  90.0               & 47.3              & 89.1            & 30.1              &70.4 \\ 
CGDM\cite{DuCVPR21}      & 93.7              &82.7               & 73.2             & 68.4               & 92.9                & 94.5              &88.7                & 82.1             &93.4                & 82.5                & 86.8                & 49.2               & 82.3 \\ 
SHOT\cite{Liang20}          &94.3               & 88.5               & 80.1                 & 57.3              &  93.1             &  93.1              & 80.7               & 80.3              & 91.5               &89.1             &  86.3           & 58.2            & 82.9 \\ 
\hline
ViT-B\cite{Dosovitskiy20}         & 97.7              & 48.1                & 86.6              & 61.6               & 78.1              &  63.4               & 94.7                & 10.3               & 87.7            &   47.7                & 94.4                &  35.5              & 67.1 \\
TVT-B\cite{Yang23}         &   92.9           & 85.6               &77.5               & 60.5              &  93.6             & 98.2            &89.4                & 76.4               & 93.6               & 92.0                & 91.7                & 55.7               & 83.9 \\ 
CDTrans-B\cite{Xu22}     & 97.1                & 90.5               &82.4              & 77.5               & 96.6             & 96.1               &  93.6                &{\color{red}\textbf{88.6}}               &  {\color{red}\textbf{97.9}}                & 86.9               &  90.3             & 62.8            & 88.4 \\ 
SSRT-B \cite{Sun2022CVPR}       & {\color{red}\textbf{98.9}}               & 87.6              & {\color{red} \textbf{89.1} }         &{\color{red} \textbf{84.8} }     & 98.3              & {\color{red}\textbf{98.7} }              &{\color{red}\textbf{96.3} }               & 81.1               &94.9             & 97.9        & 94.5                & 43.1               & 88.8 \\
NoisyTVT-B &  98.8   &  {\color{red}\textbf{95.5}}     &84.8    &  73.7     &  {\color{red}\textbf{98.5}}                & 97.2     & 95.1    & 76.5    & 95.9   &  {\color{red}\textbf{98.4}}   &  {\color{red}\textbf{98.3}}                &  {\color{red}\textbf{67.2}}    & {\color{red}\textbf{90.0}} \\ \hline
\end{tabular}
\label{Visda17Appendix}
\end{table*}

\subsection{Positive Noise for Domain Adaptation}
Unsupervised domain adaptation (UDA) aims to learn transferable knowledge across the source and target domains with different distributions \cite{Pan09Survey} \cite{Ying18Transfer}. There are mainly two kinds of deep neural networks for UDA, which are CNN-based and Transformer-based methods \cite{Sun2022CVPR} \cite{Yang23}. Various techniques for UDA are adopted on these backbone architectures. For example, the discrepancy techniques measure the distribution divergence between source and target domains \cite{Long18} \cite{Sun18Coral}. Adversarial adaptation discriminates domain-invariant and domain-specific representations by playing an adversarial game between the feature extractor and a domain discriminator \cite{Ganin15}.

\noindent Recently, transformer-based methods achieved SOTA results on UDA, therefore, we evaluate the ViT-B with the positive noise on widely used UDA benchmarks. Here the positive noise is the linear transform noise identical to that used in the classification task. The positive noise is injected into the last layer of the model, the same as the classification task. The datasets include \textbf{Office Home} \cite{Venkateswara17} and \textbf{VisDA2017} \cite{Peng17}. \textbf{Office-Home}\cite{Venkateswara17} has 15,500 images of 65 classes from four domains: Artistic (Ar), Clip Art (Cl), Product (Pr), and Real-world (Rw) images. \textbf{VisDA2017} is a Synthetic-to-Real object recognition dataset, with more than 0.2 million images in 12 classes. We use the ViT-B with a $16\times 16$ patch size, pre-trained on ImageNet. We use minibatch Stochastic Gradient Descent (SGD) optimizer \cite{Ruder18} with a momentum of 0.9 as the optimizer. The batch size is set to 32. We initialized the learning rate as 0 and linearly warm up to 0.05 after 500 training steps. The results are shown in Table \ref{Office-HomeAppendix} and \ref{Visda17Appendix}. The methods above the black line are based on CNN architecture, while those under the black line are developed from the Transformer architecture. The NoisyTVT-B, i.e., TVT-B with positive noise, achieves better performance than existing works. These results show that positive noise can improve model generality and, therefore, benefit deep models in domain adaptation tasks.

%\footnote{As in this example.}
\end{appendices}

\bibliographystyle{plain}
\bibliography{PN}

\begin{thebibliography}{10}

\bibitem{Shaykh98}
Osama~K. Al-Shaykh and Russell~M. Mersereau.
\newblock Lossy compression of noisy images.
\newblock {\em IEEE Transactions on Image Processing}, 7(12):1641--1652, 1998.

\bibitem{Albukhanajer2014}
Wissam~A. Albukhanajer, Johann~A. Briffa, and Yaochu Jin.
\newblock Evolutionary multiobjective image feature extraction in the presence of noise.
\newblock {\em IEEE Transactions on Cybernetics}, 45(9):1757--1768, 2014.

\bibitem{beit}
Hangbo Bao, Li~Dong, and Furu Wei.
\newblock {BEiT}: {BERT} pre-training of image transformers.
\newblock {\em arXiv preprint arXiv:2106.08254}, 2021.

\bibitem{Benzi1981}
Roberto Benzi, Alfonso Sutera, and Angelo Vulpiani.
\newblock The mechanism of stochastic resonance.
\newblock {\em Journal of Physics A: mathematical and general}, 14(11):L453, 1981.

\bibitem{Box64}
George~EP Box and David~R. Cox.
\newblock An analysis of transformations.
\newblock {\em Journal of the Royal Statistical Society: Series B (Methodological)}, 26(2):211--243, 1964.

\bibitem{Braun2021}
Sebastian Braun, Hannes Gamper, Chandan~KA Reddy, and Ivan Tashev.
\newblock Towards efficient models for real-time deep noise suppression.
\newblock In {\em ICASSP 2021-2021 IEEE International Conference on Acoustics, Speech and Signal Processing (ICASSP)}, pages 656--660, 2021.

\bibitem{Chan05}
Raymond~H. Chan, Chung-Wa Ho, and Mila Nikolova.
\newblock Salt-and-pepper noise removal by median-type noise detectors and detail-preserving regularization.
\newblock {\em IEEE Transactions on image processing}, 14(10):1479--1485, 2005.

\bibitem{Chan2005}
Raymond~H. Chan, Chung-Wa Ho, and Mila Nikolova.
\newblock Salt-and-pepper noise removal by median-type noise detectors and detail-preserving regularization.
\newblock {\em IEEE Transactions on image processing}, 14(10):1479--1485, 2005.

\bibitem{Cover99}
Thomas~M. Cover.
\newblock Elements of information theory.
\newblock {\em John Wiley \& Sons}, 1999.

\bibitem{CuiCVPR20}
Shuhao Cui, Shuhui Wang, Junbao Zhuo, Liang Li, Qingming Huang, and Qi~Tian.
\newblock Towards discriminability and diversity: Batch nuclear-norm maximization under label insufficient situations.
\newblock {\em CVPR}, pages 3941--3950, 2020.

\bibitem{d2021convit}
St{\'e}phane d'Ascoli, Hugo Touvron, Matthew Leavitt, Ari Morcos, Giulio Biroli, and Levent Sagun.
\newblock Convit: Improving vision transformers with soft convolutional inductive biases.
\newblock {\em arXiv preprint arXiv:2103.10697}, 2021.

\bibitem{Dehghani2023}
Mostafa Dehghani, Josip Djolonga, Basil Mustafa, Piotr Padlewski, Jonathan Heek, Justin Gilmer, Andreas Steiner, and et~al.
\newblock Scaling vision transformers to 22 billion parameters.
\newblock {\em arXiv preprint arXiv:2302.05442 (2023)}, 2023.

\bibitem{Deng2009}
Jia Deng, Wei Dong, Richard Socher, Li-Jia Li, Kai Li, and Feifei Li.
\newblock Imagenet: A large-scale hierarchical image database.
\newblock In {\em IEEE conference on computer vision and pattern recognition}, pages 248--255, 2009.

\bibitem{DingDavit22}
Mingyu Ding, Bin Xiao, Noel Codella, Ping Luo, Jindong Wang, and Lu~Yuan.
\newblock Davit: Dual attention vision transformers.
\newblock In {\em In Computer Vision–ECCV 2022: 17th European Conference}, pages 74--92, 2022.

\bibitem{Dosovitskiy20}
Alexey Dosovitskiy, Lucas Beyer, Alexander Kolesnikov, Dirk Weissenborn, Xiaohua Zhai, Thomas Unterthiner, Mostafa Dehghani, Matthias Minderer, Georg Heigold, Sylvain Gelly, Jakob Uszkoreit, and Neil Houlsby.
\newblock An image is worth 16x16 words: Transformers for image recognition at scale.
\newblock In {\em arXiv preprint arXiv:2010.11929}, 2020.

\bibitem{DuCVPR21}
Zhekai Du, Jingjing Li, Hongzu Su, Lei Zhu, and Ke~Lu.
\newblock Cross-domain gradient discrepancy minimization for unsupervised domain adaptation.
\newblock {\em CVPR}, pages 3937--3946, 2021.

\bibitem{Feng2014}
Changyong Feng, Hongyue Wang, Naiji Lu, Tian Chen, Hua He, Ying Lu, and Xin~M. Tu.
\newblock Log-transformation and its implications for data analysis.
\newblock {\em Shanghai archives of psychiatry}, 26(2):105, 2014.

\bibitem{Frankel11Kronecker}
Theodore Frankel.
\newblock {\em The geometry of physics: an introduction}.
\newblock Cambridge university press, Cambridge, 2011.

\bibitem{Ganin15}
Yaroslav Ganin and Victor Lempitsky.
\newblock Unsupervised domain adaptation by backpropagation.
\newblock {\em ICML}, pages 1180--1189, 2015.

\bibitem{Gonzales2007}
Rafael~C. Gonzales and Paul Wintz.
\newblock {\em Digital image processing}.
\newblock Addison-Wesley Longman Publishing Co., Inc., 1987.

\bibitem{Grandvalet04}
Yves Grandvalet and Yoshua Bengio.
\newblock Semi-supervised learning by entropy minimization.
\newblock {\em NIPS}, pages 211--252, 2004.

\bibitem{Kaiming16}
Kaiming He, Xiangyu Zhang, Shaoqing Ren, and Jian Sun.
\newblock Deep residual learning for image recognition.
\newblock In {\em Proceedings of the IEEE conference on computer vision and pattern recognition}, pages 770--778, 2016.

\bibitem{Horn2012}
Roger~A. Horn and Johnson~Charles R.
\newblock {\em Matrix analysis}.
\newblock Cambridge university press, 2012.

\bibitem{Johnson1995}
Norman~L. Johnson, Samuel Kotz, and Narayanaswamy Balakrishnan.
\newblock {\em Continuous univariate distributions, volume 2}.
\newblock John wiley \& sons, 1995.

\bibitem{Kannan1979}
Ravindran Kannan and Achim Bachem.
\newblock Polynomial algorithms for computing the smith and hermite normal forms of an integer matrix.
\newblock {\em siam Journal on Computing}, 8(4):499--507, 1979.

\bibitem{Kullback51}
Solomon Kullback and Richard~A. Leibler.
\newblock On information and sufficiency.
\newblock {\em The annals of mathematical statistics}, 22(1):79--86, 1951.

\bibitem{Le2015}
Ya~Le and Xuan Yang.
\newblock Tiny imagenet visual recognition challenge.
\newblock {\em CS 231N 7}, (7), 2015.

\bibitem{LeCun95}
Yann LeCun and Yoshua Bengio.
\newblock Convolutional networks for images, speech, and time series.
\newblock {\em The handbook of brain theory and neural networks}, 3361(10), 1995.

\bibitem{LiAAAI20}
Shuang Li, Chi Liu, Qiuxia Lin, Binhui Xie, Zhengming Ding, Gao Huang, and Jian Tang.
\newblock Domain conditioned adaptation network.
\newblock {\em AAAI}, pages 11386--11393, 2020.

\bibitem{Li2022Positive}
Xuelong Li.
\newblock Positive-incentive noise.
\newblock {\em IEEE Transactions on Neural Networks and Learning Systems}, 2022.

\bibitem{Liang20}
Jian Liang, Dapeng Hu, and Jiashi Feng.
\newblock Do we really need to access the source data? source hypothesis transfer for unsupervised domain adaptation.
\newblock {\em ICML}, pages 6028--6039, 2020.

\bibitem{LiangCVPR21}
Jian Liang, Dapeng Hu, and Jiashi Feng.
\newblock Domain adaptation with auxiliary target domain-oriented classifier.
\newblock {\em CVPR}, pages 16632--16642, 2021.

\bibitem{liu2021Swin}
Ze~Liu, Yutong Lin, Yue Cao, Han Hu, Yixuan Wei, Zheng Zhang, Stephen Lin, and Baining Guo.
\newblock Swin transformer: Hierarchical vision transformer using shifted windows.
\newblock In {\em Proceedings of the IEEE/CVF International Conference on Computer Vision (ICCV)}, 2021.

\bibitem{Long18}
Mingsheng Long, Zhangjie Cao, Jianmin Wang, and Michael Jordan.
\newblock Conditional adversarial domain adaptation.
\newblock In {\em Advances in neural information processing systems}, pages 1645--1655, 2018.

\bibitem{Loshchilov2017}
Ilya Loshchilov and Frank Hutter.
\newblock Decoupled weight decay regularization.
\newblock {\em arXiv preprint arXiv:1711.05101}, 2017.

\bibitem{Marcus1990}
Marvin Marcus.
\newblock Determinants of sums.
\newblock {\em The College Mathematics Journal}, 2:130--135, 1990.

\bibitem{McClintock2002}
Peter McClintock.
\newblock Can noise actually boost brain power?
\newblock {\em Physics World}, 15(7), 2002.

\bibitem{Mood1950}
Alexander~McFarlane Mood.
\newblock {\em Introduction to the Theory of Statistics}.
\newblock 1950.

\bibitem{Mori2002}
Toshio Mori and Shoichi Kai.
\newblock Noise-induced entrainment and stochastic resonance in human brain waves.
\newblock {\em Physical review letters}, 88(21), 2002.

\bibitem{Ormiston20}
Rich Ormiston, Tri Nguyen, Michael Coughlin, Rana~X. Adhikari, and Erik Katsavounidis.
\newblock Noise reduction in gravitational-wave data via deep learning.
\newblock {\em Physical Review Research}, 2(3):033066, 2020.

\bibitem{Owotogbe19}
J.~S. Owotogbe, T.~S. Ibiyemi, and B.~A. Adu.
\newblock A comprehensive review on various types of noise in image processing.
\newblock {\em int. J. Sci. eng. res}, 10(10):388--393, 2019.

\bibitem{Pan09Survey}
Sinno~Jialin Pan and Qiang Yang.
\newblock A survey on transfer learning.
\newblock {\em IEEE Transactions on knowledge and data engineering}, 22(10):1345--1359, 2009.

\bibitem{Peng17}
Xingchao Peng, Ben Usman, Neela Kaushik, Judy Hoffman, Dequan Wang, and Kate Saenko.
\newblock Visda: The visual domain adaptation challenge.
\newblock {\em arXiv preprint arXiv:1710.06924}, 2017.

\bibitem{Pereira2021}
Lis~Kanashiro Pereira, Yuki Taya, and Ichiro Kobayashi.
\newblock Multi-layer random perturbation training for improving model generalization efficiently.
\newblock {\em Proceedings of the Fourth BlackboxNLP Workshop on Analyzing and Interpreting Neural Networks for NLP}, 2021.

\bibitem{Radnosrati2020}
Kamiar Radnosrati, Gustaf Hendeby, and Fredrik Gustafsson.
\newblock Crackling noise.
\newblock {\em IEEE Transactions on Signal Processing}, 68:3590--3602, 2020.

\bibitem{Ruder18}
Sebastian Ruder.
\newblock An overview of gradient descent optimization algorithms.
\newblock {\em arXiv preprint arXiv:1609.04747}, 2016.

\bibitem{Russo03}
Fabrizio Russo.
\newblock A method for estimation and filtering of gaussian noise in images.
\newblock {\em IEEE Transactions on Instrumentation and Measurement}, 52(4):1148--1154, 2003.

\bibitem{Sethna2001}
James~P. Sethna, Karin~A. Dahmen, and Christopher~R. Myers.
\newblock Crackling noise.
\newblock {\em Nature}, 410(6825):242--250, 2001.

\bibitem{Shalev14book}
Shai Shalev-Shwartz and Shai Ben-David.
\newblock {\em Understanding machine learning: From theory to algorithms}.
\newblock Cambridge university press, Cambridge, 2014.

\bibitem{Shannon01}
Claude~Elwood Shannon.
\newblock A mathematical theory of communication.
\newblock {\em ACM SIGMOBILE mobile computing and communications review}, 5(1):3--55, 2001.

\bibitem{Sherman1949}
Jack Sherman and Winifred~J. Morrison.
\newblock Adjustment of an inverse matrix corresponding to changes in the elements of a given column or a given row of the original matrix.
\newblock {\em Annals of Mathematical Statistics}, 20, 1949.

\bibitem{Shores2007}
Thomas~S Shores.
\newblock {\em Applied linear algebra and matrix analysis}.
\newblock Springer, New York, 2007.

\bibitem{Sijbers1996}
Jan Sijbers, Paul Scheunders, Noel Bonnet, Dirk~Van Dyck, and Erik Raman.
\newblock Quantification and improvement of the signal-to-noise ratio in a magnetic resonance image acquisition procedure.
\newblock {\em Magnetic resonance imaging}, 14(10):1157--1163, 1996.

\bibitem{Steiner2021}
Andreas Steiner, Alexander Kolesnikov, Xiaohua Zhai, Ross Wightman, Jakob Uszkoreit, and Lucas Beyer.
\newblock How to train your vit? data, augmentation, and regularization in vision transformers.
\newblock In {\em arXiv preprint arXiv:2106.10270}, 2021.

\bibitem{Sun18Coral}
Baochen Sun and Kate Saenko.
\newblock Deep coral: Correlation alignment for deep domain adaptation.
\newblock {\em ECCV}, pages 443--450, 2016.

\bibitem{Sun17JFT300}
Chen Sun, Abhinav Shrivastava, Saurabh Singh, and Abhinav Gupta.
\newblock Revisiting unreasonable effectiveness of data in deep learning era.
\newblock In {\em In Proceedings of the IEEE international conference on computer vision}, pages 843--852, 2017.

\bibitem{Sun2022CVPR}
Tao Sun, Cheng Lu, Tianshuo Zhang, and Harbin Ling.
\newblock Safe self-refinement for transformer-based domain adaptation.
\newblock {\em Proceedings of the IEEE/CVF Conference on Computer Vision and Pattern Recognition}, pages 7191--7200, 2022.

\bibitem{Thulasidasan19}
Sunil Thulasidasan, Tanmoy Bhattacharya, Jeff Bilmes, Gopinath Chennupati, and Jamal Mohd-Yusof.
\newblock Combating label noise in deep learning using abstention.
\newblock In {\em arXiv preprint arXiv:1905.10964}, 2019.

\bibitem{Touvron2021DeiT}
Hugo Touvron, Matthieu Cord, Matthijs Douze, Francisco Massa, Alexandre Sablayrolles, and Hervé Jégou.
\newblock Training data-efficient image transformers \& distillation through attention.
\newblock In {\em International conference on machine learning}, pages 10347--10357, 2021.

\bibitem{TuMaxvit22}
Zhengzhong Tu, Hossein Talebi, Han Zhang, Feng Yang, Peyman Milanfar, Alan Bovik, and Yinxiao Li.
\newblock Maxvit: Multi-axis vision transformer.
\newblock In {\em In Computer Vision–ECCV 2022: 17th European Conference}, pages 459--479, 2022.

\bibitem{Vaswani17}
Ashish Vaswani, Noam Shazeer, Niki Parmar, Jakob Uszkoreit, Llion Jones, Aidan~N. Gomez, Łukasz Kaiser, and Illia Polosukhin.
\newblock Attention is all you need.
\newblock In {\em Advances in neural information processing systems}, 2017.

\bibitem{Venkateswara17}
Hemanth Venkateswara, Jose Eusebio, Shayok Chakraborty, and Sethuraman Panchanathan.
\newblock Deep hashing network for unsupervised domain adaptation.
\newblock {\em CVPR}, pages 5018--5027, 2017.

\bibitem{Ying18Transfer}
Ying Wei, Yu~Zhang, Junzhou Huang, and Qiang Yang.
\newblock Transfer learning via learning to transfer.
\newblock {\em ICML}, pages 5085--5094, 2018.

\bibitem{Woodbury1950}
M.~A. Woodbury.
\newblock Inverting modified matrices.
\newblock {\em Statistical Research Group, Memorandum Report 42}, 1950.

\bibitem{Xu19}
Ruijia Xu, Guanbin Li, Jihan Yang, and Liang Lin.
\newblock Larger norm more transferable: An adaptive feature norm approach for unsupervised domain adaptation.
\newblock {\em ICCV}, pages 1426--1435, 2019.

\bibitem{Xu22}
Tongkun Xu, Weihua Chen, Fan Wang, Hao Li, and Rong Jin.
\newblock Cdtrans: Cross-domain transformer for unsupervised domain adaptation.
\newblock {\em ICLR}, pages 520--530, 2022.

\bibitem{Yang23}
Jinyu Yang, Jingjing Liu, Ning Xu, and Junzhou Huang.
\newblock Tvt: Transferable vision transformer for unsupervised domain adaptation.
\newblock {\em WACV}, pages 520--530, 2023.

\bibitem{Zhai22}
Xiaohua Zhai, Alexander Kolesnikov, Neil Houlsby, and Lucas Beyer.
\newblock Scaling vision transformers.
\newblock In {\em Proceedings of the IEEE/CVF Conference on Computer Vision and Pattern Recognition}, pages 12104--12113, 2022.

\bibitem{Zhang2023Positive}
Hongyuan Zhang, Sida Huang, and Xuelong Li.
\newblock Variational positive-incentive noise: How noise benefits models.
\newblock {\em arXiv preprint arXiv:2306.07651}, 2023.

\end{thebibliography}

\end{document}